\documentclass[lettersize,journal]{IEEEtran}
\usepackage{cite}
\usepackage{amsmath,amssymb,amsfonts}
\usepackage{algorithmic}
\usepackage{graphicx}
\usepackage{textcomp}
\usepackage{color}
\usepackage[hidelinks]{hyperref}       

\usepackage[normalem]{ulem}
\usepackage{multirow}

\usepackage[normalem]{ulem}

\hypersetup{
colorlinks
}

\usepackage[linesnumbered,ruled,vlined]{algorithm2e}

\SetCommentSty{mycommfont}

\def\BibTeX{{\rm B\kern-.05em{\sc i\kern-.025em b}\kern-.08em
    T\kern-.1667em\lower.7ex\hbox{E}\kern-.125emX}}
\begin{document}

\SetKwComment{Comment}{/* }{ */}
\RestyleAlgo{ruled}

\title{Cross- and Intra-image Prototypical Learning for Multi-label Disease Diagnosis and Interpretation}





\author{Chong Wang, \textit{Member, IEEE}, Fengbei Liu, Yuanhong Chen, Helen Frazer, Gustavo Carneiro
\thanks{This work was supported in part by funding from the Australian Government under the Medical Research Future Fund - Grant MRFAI000090 for the Transforming Breast Cancer Screening with Artificial Intelligence (BRAIx) Project. 
(Corresponding author: Chong Wang.)}
\thanks{Chong Wang and Yuanhong Chen are with the Australian Institute for Machine Learning, University of Adelaide, SA 5000, Australia 
(e-mail: chong.wang@adelaide.edu.au; yuanhong.chen@adelaide.edu.au).}
\thanks{Fengbei Liu is with the Cornell University, New York City, NY 10044, USA
(e-mail: fl453@cornell.edu).}
\thanks{Helen Frazer is with the St Vincent’s Hospital Melbourne, VIC 3002, Australia
(e-mail: helen.frazer@svha.org.au).}
\thanks{Gustavo Carneiro is with the Centre for Vision, Speech and Signal Processing (CVSSP), University of Surrey, United Kingdom, and also with the Australian Institute for Machine Learning, University of Adelaide, SA 5000, Australia
(e-mail: g.carneiro@surrey.ac.uk).}
}

\maketitle

\begin{abstract}

Recent advances in prototypical learning have shown remarkable potential to provide useful decision interpretations associating activation maps and predictions with class-specific training prototypes.
Such prototypical learning has been well-studied for various single-label diseases, but for quite relevant and more challenging 
multi-label diagnosis, where multiple diseases are often concurrent within an image,
existing prototypical learning models struggle to obtain meaningful activation maps and effective class prototypes due to the entanglement of the multiple diseases. 
In this paper, we present a novel Cross- and Intra-image Prototypical Learning (CIPL) framework, for accurate multi-label disease diagnosis and interpretation from medical images. 
CIPL takes advantage of common cross-image semantics to disentangle the multiple diseases when learning the prototypes, 
allowing a comprehensive understanding of complicated pathological lesions. 
Furthermore, we propose a new two-level alignment-based regularisation strategy that effectively leverages consistent intra-image information to enhance interpretation robustness and predictive performance. 
Extensive experiments show that our CIPL attains the state-of-the-art (SOTA) classification accuracy in two public multi-label benchmarks of disease diagnosis: thoracic radiography and fundus images. 
Quantitative interpretability results show that CIPL also has superiority in weakly-supervised thoracic disease localisation over other leading saliency- and prototype-based explanation methods. 

\end{abstract}


\vspace{15pt}
\section{Introduction}
\label{section:introduction}

\IEEEPARstart{T}{he} last decade has witnessed the success of deep learning~\cite{lecun2015deep,wang2020automatic}, e.g., deep neural networks (DNNs), in supporting disease diagnosis and treatment in healthcare, including disease classification~\cite{ma2019multi,fang2019attention,chen2022multi,frazer2024comparison}, lesion detection~\cite{tian2022contrastive,chen2024braixdet}, and organ segmentation~\cite{fang2017automatic,wang2022bowelnet,liu2022translation}. 
Typically, DNNs require intricate network architectures, high-dimensional feature representations, and a massive number of trainable network parameters, posing significant challenges in interpreting the internal mechanisms of the model and gaining insights into how it makes predictions. 
As a result, DNNs are often considered as black-box models~\cite{rudin2019stop}.
However, in safety-critical healthcare applications with profound implications for human lives, there is a growing demand to utilise interpretable deep-learning disease diagnosis algorithms that can be effortlessly comprehended and trusted by human experts~\cite{rudin2019stop,van2022explainable}.

Most existing disease diagnosis algorithms attempt to improve model interpretability or transparency with saliency maps 
that are typically based on post-hoc explanations~\cite{patricio2023explainable} (e.g., CAM and its variants~\cite{selvaraju2017grad}, visual attentions~\cite{ouyang2020learning}, counterfactual example~\cite{singla2023explaining}, and occlusion sensitivity~\cite{shahamat2020brain}). 
However, these methods have many drawbacks, such as: 
1) saliency maps only give a spatial importance of image regions related to the model's predictions but they do not explain how those image regions are used for the predictions, 
which are not enough to interpret the model's inner working~\cite{rudin2019stop,van2022explainable}; 
2) saliency maps could be essentially the same for different classes, as evidenced in~\cite{rudin2019stop};
and 3) post-hoc explanations are often unreliable or even misleading since they are not involved in the model’s training to penalise explanation errors, 
unless additional regularisation training objectives~\cite{mahapatra2022interpretability,weber2023beyond} are applied to constrain the saliency maps.

\begin{figure*}[!t] 
\centering
\includegraphics[width=1.0\linewidth]{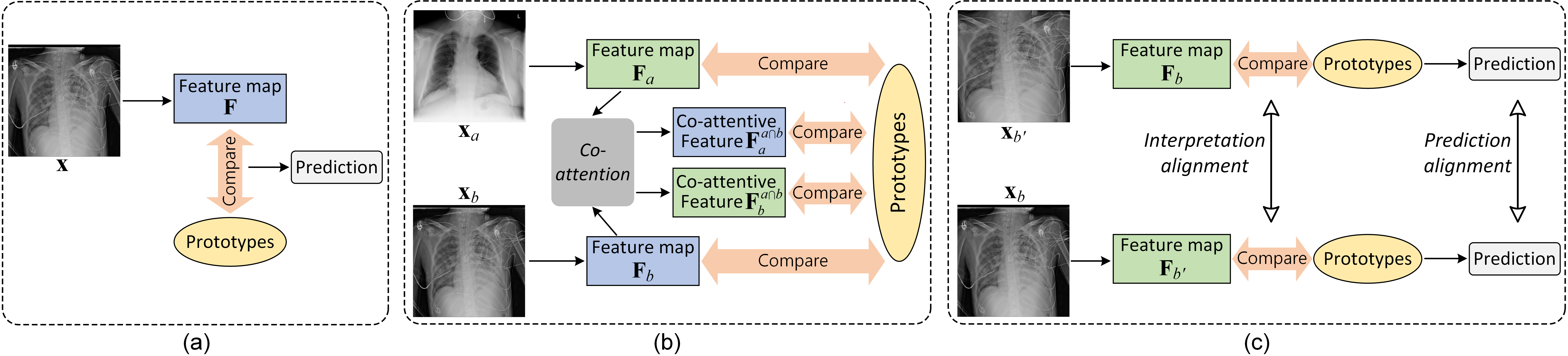}
\vspace{-20 pt}
\caption{In multi-label learning, 
(a) conventional prototypical learning strategy directly learns class prototypes from entangled multiple diseases present in training samples, by considering only individual-image information; 
(b) our cross-image prototypical learning strategy leverages common semantics of paired images to learn class prototypes from disentangled multiple diseases; 
(c) our intra-image prototypical learning strategy exploits consistent cues between paired augmented views of an image for regularising both interpretations and predictions. 
}
\label{fig:motivation}
\vspace{-10pt}
\end{figure*}

These drawbacks are mitigated with model interpretability by prototypical learning~\cite{snell2017prototypical,zhou2022rethinking}, as demonstrated in Fig.~\ref{fig:motivation}(a), which 
leverages self-explaining models~\cite{chen2019looks,donnelly2022deformable,sacha2023protoseg}.
The basic notion behind this strategy is that the model is trained to automatically 
learn 
class-specific prototypes (image-level~\cite{snell2017prototypical}, region-level~\cite{chen2019looks}, or pixel-level~\cite{sacha2023protoseg,wu2024cephalometric}) from training samples.
Then, the prediction interpretation is performed by comparing a testing sample with the learned prototypes, which aiding in interpreting the model's inner workings.
Such a self-explaining strategy relies on train-test sample associations for decision making, which resembles how humans reason according to cognitive psychological studies~\cite{aamodt1994case,yang2021multiple} revealing that humans use past cases as models when learning to solve problems. 
Because of this intuitive nature, recent studies successfully leveraged it for single-label classification problems, i.e., diagnosis and interpretation of the breast cancer~\cite{wang2022Knowledge}, diabetic retinopathy~\cite{hesse2022insightr}, and Alzheimer’s disease~\cite{wolf2023don},
where class-specific prototypes are clearly learned from the single-label training samples of the corresponding class.
However, the application of the prototype-based interpretation strategy to multi-label classification problems faces critical challenges, 
mostly due to the complexity of multi-label samples, which inherently intertwine representative disease features from multiple classes. 
Such complexity presents significant obstacles to the effective learning of the class-specific prototypes, 
resulting in prototypes that are highly entangled with multi-class disease features. 

In this paper, we address the problem of multi-label disease diagnosis and interpretation via \textbf{C}ross- and \textbf{I}ntra-image \textbf{P}rototypical \textbf{L}earning (\textbf{CIPL}). 
The overall motivation of our CIPL method is illustrated in Fig.~\ref{fig:motivation}. 
Compared with the conventional prototypical learning strategy in Fig.~\ref{fig:motivation}(a) 
that merely learns class prototypes from multi-label training images containing entangled disease features of multiple classes, 
we propose to leverage the common cross-image semantics of paired images to disentangle the multi-class disease features with a co-attention mechanism, as shown in Fig.~\ref{fig:motivation}(b).
In this way, our CIPL method enables the class prototypes to gain a comprehensive understanding of pathological lesions for each disease class.
Moreover, to obtain robust interpretation and improved classification, 
our CIPL further exploits consistent and meaningful intra-image cues between different views of an image, as demonstrated in Fig.~\ref{fig:motivation}(c).
This is achieved through a two-level alignment-based paradigm, which regularises the model interpretations and predictions in a self-supervised fashion. 
Generally, the cross-image and and intra-image strategies can complementarily contribute to the effective learning of class prototypes from multi-label training samples. 

In summary, our main contributions are listed as follows: 
\begin{enumerate}
    \item We present the CIPL method to address the multi-label disease diagnosis and interpretation with prototypical learning. 
    CIPL leverages rich cross-image semantics and consistent intra-image cues to complementarily facilitate the prototype learning from multi-label images. 
    \item We propose a novel multi-disease disentangling strategy by discovering common cross-image semantics between multi-label samples, 
    enabling the class prototypes to comprehensively understand the pathological lesions. 
    \item We introduce a new regularisation paradigm based on two-level alignments which exploit consistent intra-image information to promote both interpretation robustness and predictive performance. 
\end{enumerate}
Extensive experiments on two public multi-label benchmarks reveal that our CIPL achieves the state-of-the-art (SOTA) disease diagnosis and interpretation results.


\vspace{6pt}
\section{Related Work}
\label{sec:relatedwork}
\vspace{6pt}
\subsection{Multi-label Disease Diagnosis}
\label{sec:relatedwork_multilabel}

From a general domain like computer vision, multi-label classification (MLC) has been studied extensively in various contexts~\cite{wu2020distribution,lanchantin2021general,yan2022inferring,hang2022end}, 
e.g., handling class label imbalance~\cite{wu2020distribution}
and exploring dependencies between labels~\cite{lanchantin2021general}. 
For example, 
DB-Focal~\cite{wu2020distribution} devises a distribution-balanced loss to tackle the negative-label dominance issue in long-tailed multi-label scenarios. 
C-Tran~\cite{lanchantin2021general} adopts a convolutional neural network (CNN) encoder to extract visual features from images and a visual transformer encoder to capture complex dependencies among visual features and labels. 
CPCL~\cite{zhou2021multi} improves classification performance by exploiting the compositional nature of multi-label images through category prototypes.


In recent years, MLC has exhibited widespread use in medical imaging domain for the diagnosis of multi-label diseases, 
where one of the most common scenarios is thoracic diseases in chest X-rays~\cite{rajpurkar2017chexnet,zhou2021contrast,mahapatra2023class,hasanah2024chexnet,faisal2023chexvit,liu2022nvum,chen2023bomd}.
For instance, CheXNet~\cite{rajpurkar2017chexnet} employs a deep dense convolutional network (DenseNet121) for classifying chest radiology, 
reaching a radiologist-level pneumonia detection performance. 
MAE~\cite{xiao2023delving} further improves the performance by pre-training the DenseNet121 network with modern self-supervised techniques, e.g., masked autoencoder~\cite{he2022masked}. 
AnaXNet~\cite{agu2021anaxnet} adopts graph convolutional networks to model dependencies between disease labels as well as relationships between anatomical regions in multi-label chest X-rays. 
In~\cite{ouyang2020learning}, a hierarchical (i.e., foreground, positive, and abnormality) attention mining method is proposed for multi-label thoracic disease diagnosis and localisation from chest X-ray images. 
In~\cite{zhou2021contrast}, a multi-modal thoracic disease reasoning framework is developed by harnessing both visual features of X-ray images and graph embeddings of the disease knowledge. 
MARL~\cite{nie2023deep} presents a deep reinforcement learning framework for chest X-ray classification, enabling the diagnostic agent to be exploratory and contributing to improved classification accuracy. 
ThoraX-PriorNet~\cite{hossain2023thorax} exploits both the disease-specific prior and lung region prior as guidance to attend to meaningful anatomical regions for enhanced thorax disease classification. 
In~\cite{zhang2023triplet}, a framework called TA-DCL, which incorporates triplet attention with dual-pool contrastive learning, is designed for generalised multi-label classification in medical images, including thoracic and ophthalmic diseases. 
Recent researches~\cite{zhang2022contrastive,wu2023medklip,zhang2023knowledge,dai2024unichest} have been shifting towards developing chest X-ray foundation models based on visual-language pre-training techniques, 
demonstrating leading performances on some widely-used public benchmarks. 





Regardless of their encouraging results, the aforementioned approaches mostly rely on black-box deep-learning models, 
which produce only a diagnostic outcome with poor interpretability to human users, potentially impeding their successful translation into real-world clinical practice. 
In contrast, interpretable deep-learning models not only provide the diagnostic prediction, but also elucidate the reasoning behind their decisions. 
This transparency is highly beneficial for model debugging, as it allows for the analysis of misbehavior when incorrect or unexpected predictions occur~\cite{rudin2019stop}. 

\vspace{-5pt}
\subsection{Prototypical Learning for Interpretation}

The notion of prototypical learning can be traced back to~\cite{snell2017prototypical} where 
the seminal prototypical network is designed to learn a metric space in which classification can be made by computing similarities to prototype representations of each class.
In~\cite{snell2017prototypical}, the prototypes are image-level since they are estimated as the mean of embedded feature vectors of whole images for each class. 
Derived from~\cite{snell2017prototypical}, some follow-up studies introduce the idea of prototypical learning for model interpretation in image classification. 
One important representative is ProtoPNet~\cite{chen2019looks,wang2023learning}, which automatically learns region-level prototypes to represent object-parts from training images for each class and then realises interpretable reasoning by evaluating similarities between a testing image and the learned training prototypes (prototypical parts). 
Due to this exemplar-driven nature, some works extend to use pixel-wise prototypes for improving the transparency of semantic segmentation models~\cite{zhou2022rethinking,sacha2023protoseg}. 

Recently, the prototypical learning has been explored for single-label disease diagnosis and interpretation, e.g., 
diabetic retinopathy grading~\cite{hesse2022insightr}, 
breast cancer screening~\cite{wang2022Knowledge,wang2023interpretable}, 
and Alzheimer’s disease detection~\cite{wolf2023don}. 
In these single-label tasks, the class-specific prototypes can be readily obtained by directly learning from the single-label training samples of the corresponding class, 
to capture representative and discriminative visual patterns. 
However, 
regarding the more challenging multi-label problem, a training sample is prone to be associated with multiple labels simultaneously. 
This indicates that the discriminative features of multiple classes are entangled together, undermining the effective learning of class prototypes.
To overcome this issue, we propose a cross-image prototype learning strategy with a co-attention mechanism to disentangle the disease features of multiple classes 
so that the class prototypes can comprehensively represent the pathological lesions of each disease class. 
In addition, we also present a two-level alignment-based regularisation strategy that leverages consistent intra-image cues to obtain robust interpretation and improved classification. 

\begin{figure*}[!t] 
\centering
\includegraphics[width=1.0\linewidth]{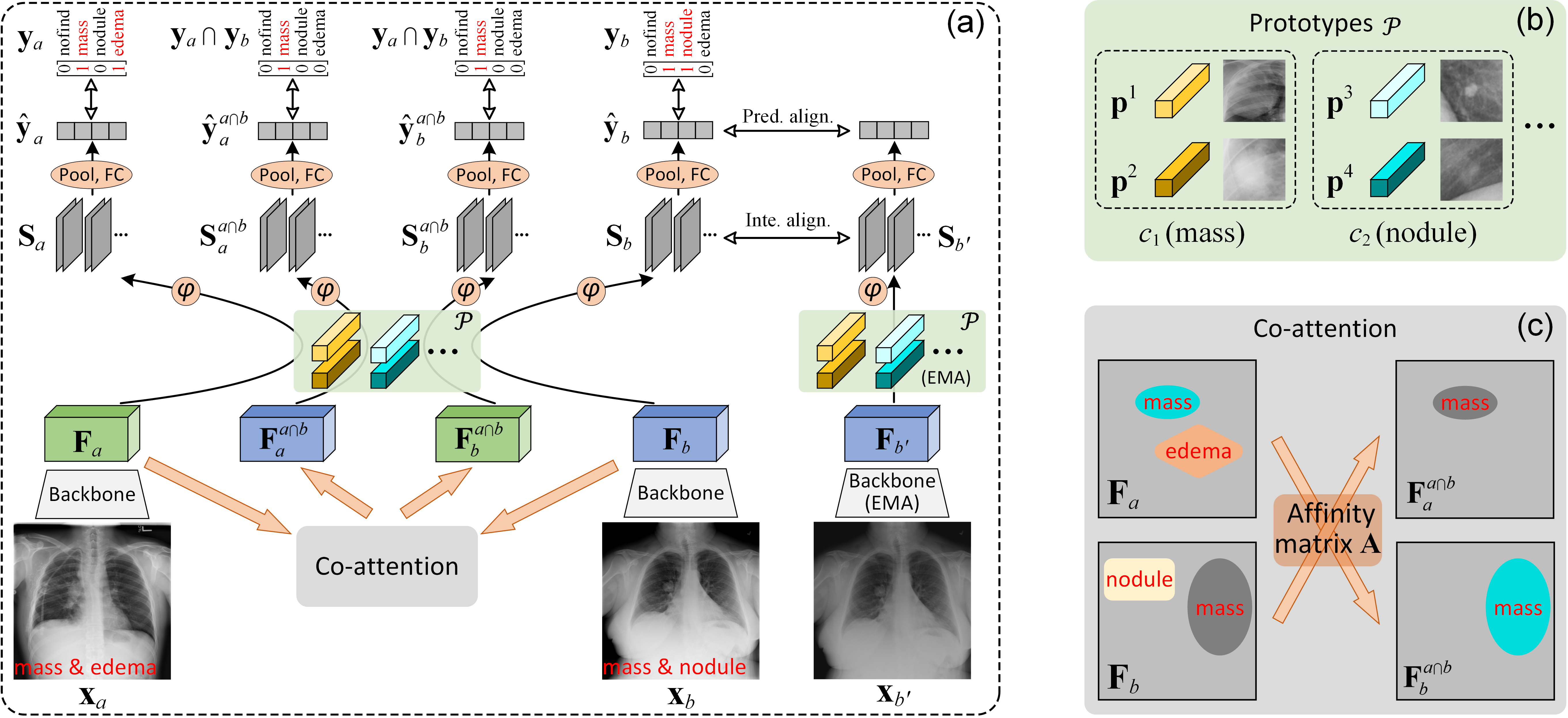}
\vspace{-18 pt}
\caption{
(a) Architecture of our proposed CIPL method for the multi-label disease diagnosis and interpretation. 
CIPL leverages a co-attention mechanism to mine cross-image semantics from paired images $\mathbf{x}_a$ and $\mathbf{x}_b$, 
with the goal of learning disentangled class prototypes from multi-label training samples. 
CIPL also regularises the prototype's learning with interpretation and prediction consistency between augmented image views, e.g., $\mathbf{x}_b$ and $\mathbf{x}_{b^{\prime}}$; 
(b) examples of prototypes that encode representative pathological lesions for each disease class (here we illustrate two prototypes per class); 
(c) co-attention mechanism extracts common semantics between paired images for multi-class disease disentangling. 
}
\label{fig:CIPL}
\vspace{-8 pt}
\end{figure*}

\section{Methodology}
\label{sec:method}

We introduce our CIPL method for diagnosis and interpretation of multi-label diseases from medical images.
Given a set of training images with image-level labels, 
CIPL is trained to automatically learn region-level prototypes for each class that encode representative and discriminative visual features, e.g. (pathological lesions or normal tissues).
Since we target the multi-label problem, CIPL further makes full use of additional cross- and intra-image information to benefit the learning of class prototypes. 
During testing, CIPL realises interpretable reasoning by evaluating similarities between a testing image and the class prototypes of training image regions.

Let us denote the training set with $\mathcal{D} = \{ (\mathbf{x}_a,\mathbf{y}_a) \}_{a=1}^{|\mathcal{D}|}$, 
where $\mathbf{x} \in \mathcal{X} \subset \mathbb{R}^{H \times W \times 3}$ is an image of size $H \times W$ and 3 colour channels, 
$\mathbf{y}_a = [y^1, y^2, ..., y^C]^{\top} \in \mathcal{Y} \subset \{0,1\}^C$ denotes the image-level class label, 
where $C$ represents the number of all possible disease labels, 
and $y^c$ is a binary indicator, i.e., $y^c$ = 1 if the disease of label $c$ is present in the image $\mathbf{x}_a$, and $y^c$ = 0 otherwise. 
As in conventional MLC, we formulate the multi-disease diagnosis problem with binary classification (one disease vs. no-findings/normal) for each of the $C$ disease labels.
In other words, our CIPL achieves interpretable multi-label predictions by comparing a testing image with prototypes of each disease class as well as those of the no-findings class. 
If there are multiple disease classes where the similarities with their respective class prototypes are larger than that of the no-findings class, 
the testing image will be classified with all these multiple disease labels.

\vspace{-5pt}
\subsection{Method Overview}
\label{sec:overview}

An overview of our CIPL is shown in Fig.~\ref{fig:CIPL}, where training image pairs, represented with $\mathbf{x}_a$, $\mathbf{x}_b$, are sampled from $\mathcal{D}$ and then fed into the same feature backbone $f_{\theta_f}:\mathcal{X} \to \mathcal{F}$ (parameterised by $\theta_f$) to extract the feature maps $\mathbf{F}_a,\mathbf{F}_b \in \mathcal{F} \subset \mathbb{R}^{\bar{H} \times \bar{W} \times D}$, with $D$ denoting the feature dimension, and usually $\bar{H} < H$, $\bar{W} < W$.
Following the general protocol of prototypical learning~\cite{chen2019looks,donnelly2022deformable,wang2022Knowledge}, 
a prototype layer $\varphi_{\mathcal{P}}:\mathcal{F} \to \mathcal{S}$ learns $N=M(C+1)$ class-specific prototypes $\mathcal{P}=\{\mathbf{p}^n\}_{n=1}^{N}$, 
with $M$ prototypes per class (including no-findings) and $\mathbf{p}^n \in \mathbb{R}^ {1 \times 1 \times {D}}$, as shown in Fig.~\ref{fig:CIPL}(b). 
The prototype layer computes similarity maps $\mathbf{S}_a$, $\mathbf{S}_b \in \mathcal{S} \subset \mathbb{R}^{\bar{H} \times \bar{W} \times N}$ between the feature maps $\mathbf{F}_a$, $\mathbf{F}_b$ and prototypes $\mathcal{P}$, e.g., $\mathbf{S}_{a}^{n, (i,j)} = \exp^{{-||\mathbf{F}_a^{(i,j)}-\mathbf{p}^n ||_2^2}\normalsize{/}{D}}$ for $\mathbf{F}_a$, 
where $i \in \{1, ..., \bar{H}\}$ and $j \in \{1, ..., \bar{W}\}$ are spatial indexes.
Due to the class-specific property of prototypes, the similarity maps $\mathbf{S}_a$, $\mathbf{S}_b$ can be viewed as class-wise prototypical activation maps, which are reduced to represent $N$ similarity scores using max-pooling, e.g., $\mathbf{s}_a^n = \max_{(i,j)} \mathbf{S}_a^{n, (i,j)}$ for $\mathbf{S}_a$.
These $N$ similarity scores for each image are fed to a fully-connected (FC) layer $g_{\theta_{g}}:\mathbb{R}^N \to \mathbb{R}^{C+1}$ to generate $C+1$ classification scores (logits).
Given the multi-label problem, we reorganise the logits to produce $C$ binary tasks, with each of them corresponding to one disease vs. no-findings.  
Finally, a softmax function is employed to derive the $C$ binary probabilistic predictions $\widehat{\mathbf{y}}_a, \widehat{\mathbf{y}}_b \in \mathbb{R}^{C \times 2}$. 
The basic training objective of our CIPL method is defined as follows:
\begin{equation}
\begin{split}
    \mathcal{L}_{basic} = \ell_{ce}(\widehat{\mathbf{y}}_a, \mathbf{y}_a) + \ell_{ce}(\widehat{\mathbf{y}}_b, \mathbf{y}_b) + \alpha_1 (\mathcal{L}_{cst} + \mathcal{L}_{sep}),
\end{split}
\label{eq:loss_basic}
\end{equation}
where $\ell_{ce}(\cdot)$ is the cross-entropy loss, $\alpha_1$ is a weighting factor for the two commonly-used auxiliary losses~\cite{chen2019looks,donnelly2022deformable,wang2022Knowledge}, i.e., 
clustering $\mathcal{L}_{cst}$ and separation $\mathcal{L}_{sep}$:
\begin{equation}
    \mathcal{L}_{cst} = 
    \min_{\mathbf{p}^n\in\mathcal{P}_{\mathbf{y}_a}} \min_{\mathbf{f} \in\mathbf{F}_a}||\mathbf{f}-\mathbf{p}^n||_2^2 + 
    \min_{\mathbf{p}^n\in\mathcal{P}_{\mathbf{y}_b}} \min_{\mathbf{f} \in\mathbf{F}_b}||\mathbf{f}-\mathbf{p}^n||_2^2,
    \label{eq:clustering}
\end{equation}
\begin{equation}
\begin{split}
    \mathcal{L}_{sep} = \ &
    \max \Big{(} 0, \tau - \min_{\mathbf{p}^n\notin\mathcal{P}_{\mathbf{y}_a}} \min_{\mathbf{f} \in\mathbf{F}_a}||\mathbf{f}-\mathbf{p}^n||_2^2 \Big{)} \ + \\
    & \max \Big{(} 0, \tau - \min_{\mathbf{p}^n\notin\mathcal{P}_{\mathbf{y}_b}} \min_{\mathbf{f} \in\mathbf{F}_b}||\mathbf{f}-\mathbf{p}^n||_2^2 \Big{)}.
\end{split}
\label{eq:seperation}
\end{equation}
Taking the training image $\mathbf{x}_a$ as an example, 
the clustering loss $\mathcal{L}_{cst}$ in Eq.~\eqref{eq:clustering} ensures the image contains at least one local feature vector $\mathbf{f} \in \mathbb{R}^ {1 \times 1 \times {D}}$ in $\mathbf{F}_a$ that is close to one of the prototypes of its own class (denoted by the set $\mathcal{P}_{\mathbf{y}_a}$),
while the separation loss $\mathcal{L}_{sep}$ in Eq.~\eqref{eq:seperation} forces all feature vectors of $\mathbf{F}_a$ to stay distant (with a positive margin $\tau$) from the prototypes that belong to other classes. 

Until now, our CIPL follows the conventional training strategy as in current approaches (e.g., ProtoPNet~\cite{chen2019looks}) that only consider individual-image information for prototype learning. 
However, a challenge arises due to the entanglement of multi-class disease features in multi-label samples, which significantly hinders the effective learning of the class prototypes. 
To handle this issue, we describe in Section~\ref{sec:cross} how our CIPL additionally mines cross-image semantics via a co-attention mechanism to facilitate the disentanglement of disease features from multiple classes, enabling the prototypes to have a more precise and comprehensive understanding of various pathological lesion patterns.
Meanwhile, we further present a two-level alignment-based strategy that regularises the prototype learning by exploiting consistent intra-image information, as elaborated in Section~\ref{sec:intra}.

\vspace{-5pt}
\subsection{Cross-image Prototypical Learning via Co-attention}
\label{sec:cross}

In multi-label diagnosis problems, label co-occurrence usually leads to the entanglement of features associated with multiple diseases, 
which is the primary obstacle for the effective learning of class-specific prototypes from multi-label samples. 
In this section, we propose a solution to decouple these entangled features with the goal of improving the learning of class prototypes in multi-label problems.
To achieve this, we leverage the well-established co-attention technique~\cite{lu2019see,lu2021segmenting,li2021group} to take advantage of cross-image information. 
The co-attention has been shown to be effective in capturing underlying relationships between different entities, 
e.g., 
vision-language navigation~\cite{wang2020active} and 
video segmentation~\cite{lu2021segmenting}.
Motivated by the general idea of co-attention mechanisms, 
our CIPL method harnesses it to discover common semantics within pairs of training images, 
which enhances the multi-disease feature disentanglement and allows the class prototypes to capture visual patterns of complicated pathological lesions.

As illustrated in Fig.~\ref{fig:CIPL}(a), our CIPL method attends to image pairs, i.e., $\mathbf{x}_a$, $\mathbf{x}_b$, in the feature space, where the two images share at lease one common label, denoted by $\mathbf{y}_a \cap \mathbf{y}_b$.
Specifically, we first transform $\mathbf{F}_a$ and $\mathbf{F}_b$ into matrices denoted by $\bar{\mathbf{F}}_a$ and $\bar{\mathbf{F}}_b \in \mathbb{R}^{\bar{H}\bar{W} \times D}$, 
and employ them to calculate an affinity matrix $\mathbf{A} \in \mathbb{R}^{\bar{H}\bar{W} \times \bar{H}\bar{W}}$, with each entry computed as follows: 
\begin{equation}
    \mathbf{A}^{i, j} = \mathsf{sim} (\bar{\mathbf{F}}^{i}_a, \bar{\mathbf{F}}^{j}_b) = - || \bar{\mathbf{F}}^{i}_a - \bar{\mathbf{F}}^{j}_b ||_2^2,
    \label{eq:affinity}
\end{equation}
where $\bar{\mathbf{F}}_a^i$ and $\bar{\mathbf{F}}_b^j$ denote a $D$-dimensional feature vector in $\bar{\mathbf{F}}_a$ and $\bar{\mathbf{F}}_b$ at position $i$ and $j$, respectively. 
It is worth mentioning that our CIPL utilises the negative $L_2$ distance as the similarity metric $\mathsf{sim}(\cdot)$, 
since it is also used to measure the similarity between the feature maps and prototypes (see Section~\ref{sec:overview}). 
This differs from the standard co-attention strategy~\cite{lu2019see,lu2021segmenting} demanding an additional trainable parameter in computing the affinity matrix.
Based on Eq.~(\ref{eq:affinity}), we can apply a column-wise softmax operation on $\mathbf{A}$ to obtain the co-attention weights:
\begin{equation}
    \mathbf{W}_a = {\rm{softmax}} (\mathbf{A}) \in [0, 1]^{\bar{H}\bar{W} \times \bar{H}\bar{W}},
    \label{eq:attention_a}
\end{equation}
\begin{equation}
    \mathbf{W}_b = {\rm{softmax}} (\mathbf{A}^{{\top}}) \in [0, 1]^{\bar{H}\bar{W} \times \bar{H}\bar{W}}.
    \label{eq:attention_b}
\end{equation}
Next, the attention summary for the feature representation $\bar{\mathbf{F}}_a$ (or $\bar{\mathbf{F}}_b$) with respect to $\bar{\mathbf{F}}_b$ (or $\bar{\mathbf{F}}_a$) are given by:
\begin{equation}
    \bar{\mathbf{F}}^{a \cap b}_a = \bar{\mathbf{F}}_b \mathbf{W}_b \in \mathbb{R}^{D \times \bar{H}\bar{W}},
    \label{eq:attentionfeat_a}
\end{equation}
\begin{equation}
    \bar{\mathbf{F}}^{a \cap b}_b = \bar{\mathbf{F}}_a \mathbf{W}_a \in \mathbb{R}^{D \times \bar{H}\bar{W}}.
    \label{eq:attentionfeat_b}
\end{equation}
The above two co-attentive features are then reshaped to restore their spatial dimensions, i.e., ${\mathbf{F}}^{a \cap b}_a$ and ${\mathbf{F}}^{a \cap b}_b \in \mathbb{R}^{\bar{H} \times \bar{W} \times D}$, 
which are supposed to preserve the common semantics between $\mathbf{F}_a$ and $\mathbf{F}_b$ and focus on the shared object regions (e.g., pathological lesions of the same disease class) in both $\mathbf{x}_a$ and $\mathbf{x}_b$. 
In this way, the class-aware disentangled features can be favourably extracted (see Fig.~\ref{fig:CIPL}(c)), benefiting the learning of class prototypes. 
Therefore, the co-attentive features are incorporated into the prototype layer $\varphi_{\mathcal{P}}(\cdot)$ by comparing with the prototypes in $\mathcal{P}$, 
whose prediction outputs are supervised with the common labels $\mathbf{y}_a \cap \mathbf{y}_b$: 
\begin{equation}
\begin{split}
    \mathcal{L}_{cross} = \ & \ell_{ce}\big(g_{\theta_g}(\varphi_{\mathcal{P}}(\mathbf{F}^{a \cap b}_a)), \mathbf{y}_a \cap \mathbf{y}_b\big) \ + \\
    & \ell_{ce}\big(g_{\theta_g}(\varphi_{\mathcal{P}}(\mathbf{F}^{a \cap b}_b)), \mathbf{y}_a \cap \mathbf{y}_b\big).
\end{split}
\label{eq:loss_coattention}
\end{equation}

Consider the illustration in Fig.~\ref{fig:CIPL}, where \textit{mass} and \textit{edema} exist in $\mathbf{x}_a$, while \textit{mass} and \textit{nodule} are present in $\mathbf{x}_b$. 
Since the co-attention is essentially performed among all the feature position pairs between $\mathbf{x}_a$ and $\mathbf{x}_b$,
only the disease with common semantics, \textit{mass}, is preserved in the co-attentive features ${\mathbf{F}}^{a \cap b}_a$ and ${\mathbf{F}}^{a \cap b}_b$.
When using the common label, \textit{mass}, as the supervisory signal, 
the \textit{mass} prototypes will be uniquely associated with the corresponding disentangled features of \textit{mass} regions 
to comprehensively and accurately understand the pathological \textit{mass} lesions. 
From an another viewpoint, 
the co-attention across the two related images enables our CIPL to become aware that 
the semantics preserved in $\mathbf{x}_{a}$ and $\mathbf{x}_{b}$ are relevant and can be exploited to jointly learn effective prototypes for identifying the \textit{mass} disease. 
It essentially takes advantage of the context across training data. 

Apart from the co-attention based prototypical learning strategy that exploits the relevant cross-image semantics, 
we further develop an alignment-based prototypical learning paradigm that mines consistent and useful intra-image cues between different views of a training image, 
introduced in the following Section~\ref{sec:intra}.
The cross-image and intra-image strategies provide a complementary way to help the model to obtain effective prototypes from multi-label samples.






\vspace{-8pt}
\subsection{Intra-image Prototypical Learning via Alignment}
\label{sec:intra}

Inherited from the strengths of existing prototype-based approaches, 
our proposed CIPL method not only offers classification predictions for disease diagnosis but also furnishes a promising interpretable reasoning through the presentation of a set of similarity maps.
These maps are associated with disease-representative prototypes derived from training images, providing a transparent process of disease interpretation. 
In order to promote interpretation robustness and predictive performance, 
we present a self-supervised alignment-based method which regularises both the prototype learning and model predictions 
by exploiting consistent and meaningful intra-image information between image views.

In line with popular self-supervised learning techniques~\cite{jing2020self}, 
we adopt two different views, denoted by $\mathbf{x}_b, \mathbf{x}_{b^{\prime}}$, augmented from the same training image, as demonstrated in Fig.~\ref{fig:CIPL}(a).
Since our CIPL achieves model interpretations through the similarity maps with training prototypes, 
we propose to leverage the intra-image cues by enforcing interpretation consistency between the two input views. 
This is formulated as the spatial alignment between the similarity maps $\mathbf{S}_{b}$ and $\mathbf{S}_{b^{\prime}}$ from the two image views, as follows:
\begin{equation}
    \mathcal{L}_{inte}(\mathbf{x}_b,\mathbf{x}_{b^{\prime}}) = - \frac{1}{\bar{H}\bar{W}}\sum_{{(i, j)} \in \bar{H} \times \bar{W}} h(\mathbf{S}_{b}^{:, (i,j)}, \mathbf{S}_{b^{\prime}}^{:, (i,j)}),
\label{eq:loss_intraspatial}
\end{equation}
where $h(\cdot)$ computes the cosine similarity. 
In Eq.~(\ref{eq:loss_intraspatial}), our CIPL is driven to produce robust prototypes to pathological lesions characterised by varying structures and appearances. 

In addition to the interpretation alignment, we also propose to enhance alignment with respect to the classification predictions between the two image views, 
where one straightforward solution is to minimise KL-divergence of probabilistic predictions from the two views, for each individual training sample. 
However, we experimentally notice that such a solution does not lead to a notable improvement in predictive performance (detailed in our experiments). 
Consequently, we design a new batch-level alignment strategy by further considering sample relationships, which is formulated as:
\begin{equation}
    \mathcal{L}_{pred}(\mathcal{B},\mathcal{B}^{\prime}) = \frac{1}{C \cdot |\mathcal{B}|^2} \sum_{c=1}^{C} || \mathbf{G}^c-\mathbf{G}^{c, \prime} ||_2^2,
\label{eq:loss_intraglobal}
\end{equation}
where $\mathcal{B}$ and $\mathcal{B}'$ denote two independent mini-batches with size
$|\mathcal{B}| = |\mathcal{B}^{\prime}|$, 
$\mathbf{G}^c$ and $\mathbf{G}^{c, \prime} \in \mathbb{R}^{|\mathcal{B}| \times |\mathcal{B}|}$ represent the class-wise Gram matrix that measures the prediction correlations between samples within mini-batches $\mathcal{B}$ and $\mathcal{B}^{\prime}$, respectively, which are given by: 
\begin{equation}
    \mathbf{G}^c = \widehat{\mathbf{Y}}^c \cdot {(\widehat{\mathbf{Y}}^c)}^{\top}, \mathbf{G}^{c, \prime} = \widehat{\mathbf{Y}}^{c, \prime} \cdot {(\widehat{\mathbf{Y}}^{c, \prime})}^{\top},
\label{eq:gram}
\end{equation}
with 
$\widehat{\mathbf{Y}} = [\widehat{\mathbf{y}}_1, \widehat{\mathbf{y}}_2, ..., \widehat{\mathbf{y}}_{|\mathcal{B}|}],
\widehat{\mathbf{Y}}^{\prime} = [\widehat{\mathbf{y}}_1^{\prime}, \widehat{\mathbf{y}}_2^{\prime}, ..., \widehat{\mathbf{y}}_{|\mathcal{B}^{\prime}|}] \in \mathbb{R}^{|\mathcal{B}| \times C \times 2}$
denoting probabilistic predictions of all training samples of mini-batches $\mathcal{B}$ and $\mathcal{B}^{\prime}$, respectively. 
Note that in Eq.~(\ref{eq:loss_intraglobal}) and (\ref{eq:gram}) we omit notations $b, b^{\prime}$, 
defining different views of the same image, to keep definitions uncluttered. 

In order to enable a stable training, we employ the well-known momentum update technique~\cite{he2020momentum} to form an exponential moving average (EMA) model, 
$\theta^{t+1}_{\rm ema} := \lambda \theta^{t}_{\rm ema} + (1 - \lambda) \theta^{t}$, where $\lambda$ = 0.999 is the momentum coefficient. 
During training, one image view (such as $\mathbf{x}_{b}$) is fed to the current model, and another view $\mathbf{x}_{b^{\prime}}$ to the EMA model. 
The EMA model can be discarded once the training completes. 







\vspace{-8pt}
\subsection{Overall Training Objectives}

We have introduced all training components of our CIPL method for multi-label disease diagnosis and interpretation.
The overall training objective is expressed as follows:
\begin{equation}
    \mathcal{L}_{total} = \mathcal{L}_{basic} + 
    \alpha_2 \mathcal{L}_{cross} + 
    \alpha_3 \mathcal{L}_{inte} + 
    \alpha_4 \mathcal{L}_{pred},
\label{eq:total_loss}
\end{equation}
where $\alpha_2, \alpha_3$, and $\alpha_4$ are trade-off factors for the corresponding loss terms. 

To enable all prototypes in $\mathcal{P}$ to be represented by actual training image patches, 
we replace each prototype $\mathbf{p}^n$ with the nearest training patch feature of the corresponding class~\cite{chen2019looks,donnelly2022deformable}, after each training epoch, with:
\begin{equation}
     \mathbf{p}^n \leftarrow \arg\min_{\mathbf{f} \in \mathbf{F}_{a\in\{1,2,...,|\mathcal{D}|\}}}||\mathbf{f} - \mathbf{p}^n||_2^2.
\label{eq:prototype_update}
\end{equation}
This step ensures the transparency of the prototypes (i.e., grounding them in the human-understandable image space), 
so that the case-based interpretability is achieved by classifying testing images based on their similarities to the training prototypes of image patches. 
To facilitate the richness and diversity of visual patterns captured by the prototypes, in Eq.~(\ref{eq:prototype_update}) we adopt the greedy prototype replacement strategy~\cite{wang2022Knowledge},
guaranteeing that all prototypes are derived from entirely distinctive (non-repetitive) training images.
To be specific, if duplicate prototypes are observed, the next most-similar training image patch is used in the replacement instead.
Additionally, Eq.~(\ref{eq:prototype_update}) is also employed to visualise the learned class prototypes (see Fig.~\ref{fig:prototypes} in Section~\ref{sec:experiment}), as in~\cite{chen2019looks,donnelly2022deformable}.

Compared with the conventional prototypical learning approaches~\cite{chen2019looks,donnelly2022deformable,wang2022Knowledge} designed for single-label problems, 
our CIPL method does not introduce any excessive learnable network parameters. 
During the training phase, CIPL incurs a moderate increase in computational overhead due to computation of the affinity matrix, 
but it does not 
require additional computational resources in the test phase.

\vspace{-5pt}
\section{Experiments}
\label{sec:experiment}

\vspace{-5 pt}
\subsection{Data Sets}
\label{sec:dataset}

In this work, we utilise two public benchmark datasets to validate the effectiveness of our proposed CIPL method for multi-label diagnosis and interpretation, 
namely the
thoracic diseases on NIH ChestX-ray14~\cite{wang2017chestx} and 
ophthalmic diseases on Ocular Disease Intelligent Recognition (ODIR)~\cite{li2021benchmark}.

The multi-label NIH ChestX-ray14 dataset~\cite{wang2017chestx} consists of  112,120 chest X-ray images featuring 14 labels of thoracic diseases as well as a ``no-findings" label, acquired from 30,805 patients. 
In ChestX-ray14, 880 images are annotated with 984 bounding boxes, specifying the disease regions across 8 disease categories.
In our experiment, the original X-rays are resized to $H$ = $W$ = 512,
and we use ChestX-ray14 for both classification and weakly-supervised localisation of thoracic diseases. 
\textit{Disease Classification:} 
our experimental setup strictly adheres to the official train/test split in~\cite{wang2017chestx}. 
The classification results are evaluated using the AUC, F1 score, and accuracy, 
where we follow the previous studies~\cite{wu2023medklip,zhang2023knowledge} to determine the thresholds in calculating the mean F1 score and accuracy. 
\textit{Weakly-supervised Disease localisation:} 
we average the class activation maps and produce a binary localisation mask with a threshold 0.5 for a disease class.  
In alignment with previous localisation approaches~\cite{ouyang2020learning,hermoza2020region} that rely on only image-level training labels, 
our method is trained on all 111,240 images without any box annotations. 
Testing is then performed on the remaining 880 images that have box annotations. 
The localisation performance is assessed using the thresholded IoU accuracy~\cite{ouyang2020learning}, 
defined as the ratio of correctly localised cases to the total cases for each disease class. 
Correct localisation is determined by the IoU score between the predicted and ground-truth localisation masks, which should exceed a predefined IoU threshold \textit{T}(IoU).

ODIR~\cite{li2021benchmark} is a public database of colour fundus images containing multi-label ophthalmic diseases.
In ODIR, there are a total of 10,000 images from left and right eyes of 5,000 patients. 
The labels are provided for each patient, including normal, diabetes, glaucoma, cataract, aged-macular degeneration (AMD), hypertension, myopia, and other diseases. 
ODIR also offers detailed diagnostic keywords related to each fundus image, allowing to relabel the images and perform an eye-level classification. 
In this work, we employ the relabelling procedure in~\cite{xie2023fundus} to realise eye-level diagnosis. 
Our experiments follow the official train/test split~\cite{li2021benchmark}, 
where the whole ODIR is divided into train set (6970 images), on-site test set (1972 images), and off-site test set (991 images), 
after excluding low-quality images (e.g., lens dust, image offset, and invisible optic disk), as in~\cite{xie2023fundus,li2021benchmark}. 
Original fundus images undergo an automatic circular border cropping process, followed by resizing to $H = W$ = 640 as model input. 
We report the AUC, F1 score, and accuracy metrics to evaluate the classification performance.  

\begin{table*}[]
\centering
\caption{Quantitative Thoracic Disease Classification Results on NIH ChestX-ray14, Reported with the AUC on the 14 Diseases, Mean AUC (mAUC), Mean F1 (mF1), and Mean Accuracy (mACC) Metrics. Top-part Illustrates General Classification Methods and Bottom Part Illustrates Prototype-based Approaches, with Best Results Highlighted in Bold. * We Re-train the Model on Official Data Split for Fair Comparison, Using the Author's Officially Released Algorithms. }
\label{tab:xrayscls}
\setlength{\tabcolsep}{1.0 mm}
\resizebox{1.0\linewidth}{!}{

\begin{tabular}{lcccccccccccccc|c|c|c} 
\hline\hline
Method                                              & Atel  & Card  & Effu  & Infi  & Mass  & Nodu  & Pnea  & Pnex  & Cons  & Edem  & Emph  & Fibr  & P.T.  & Hern  & mAUC   & mF1 & mACC \\ 
\hline
Wang \textit{et al.}~\cite{wang2017chestx}    & 0.700 & 0.810 & 0.759 & 0.661 & 0.693 & 0.669 & 0.658 & 0.799 & 0.703 & 0.805 & 0.833 & 0.786 & 0.684 & 0.872 & 0.745  & \multirow{9}{*}{n/a} & \multirow{9}{*}{n/a} \\
Guendel \textit{et al.}~\cite{guendel2019learning}  & 0.772 & 0.881 & 0.826 & 0.710 & 0.824 & 0.756 & 0.728 & 0.851 & 0.751 & 0.841 & 0.897 & 0.819 & 0.757 & 0.901 & 0.807  & & \\
GREN~\cite{qi2022gren}                              & 0.776 & 0.882 & \textbf{0.851} & 0.660 & 0.813 & 0.721 & 0.675 & 0.873 & \textbf{0.790} & \textbf{0.893} & 0.920 & 0.795 & 0.790 & 0.867 & 0.807 &  & \\
ConVIRT~\cite{zhang2022contrastive}                 & 0.771 & 0.867 & 0.825 & 0.703 & 0.818 & 0.761 & 0.722 & 0.857 & 0.747 & 0.854 & 0.901 & 0.809 & 0.771 & 0.909  & 0.808 & & \\
CheXNet~\cite{rajpurkar2017chexnet}                 & 0.764 & 0.867 & 0.819 & 0.700 & 0.824 & 0.775 & 0.710 & 0.867 & 0.746 & 0.854 & 0.922 & 0.821 & 0.781 & 0.921 & 0.814  & & \\
MedKLIP~\cite{wu2023medklip}                        & 0.777 & 0.885 & 0.827 & 0.690 & 0.829 & 0.771 & 0.733 & 0.852 & 0.748 & 0.845 & 0.904 & 0.821 & 0.780 & 0.940 & 0.815  & & \\
Guan \textit{et al.}~\cite{guan2020multi}           & 0.781 & 0.883 & 0.831 & 0.697 & 0.830 & 0.764 & 0.725 & 0.866 & 0.758 & 0.853 & 0.911 & 0.826 & 0.780 & 0.918 & 0.816  & &  \\
Ma \textit{et al.}~\cite{ma2019multi}               & 0.777 & 0.894 & 0.829 & 0.696 & \textbf{0.838} & 0.771 & 0.722 & 0.862 & 0.750 & 0.846 & 0.908 & 0.827 & 0.779 & 0.934 & 0.817  & &  \\
AnaXNet~\cite{agu2021anaxnet}                       & 0.774 & 0.889 & 0.824 & 0.709 & 0.825 & 0.776 & 0.712 & 0.870 & 0.751 & 0.855 & 0.934 & 0.820 & 0.788 & 0.924 & 0.818  & & \\
CPCL~\cite{zhou2021multi}                 & 0.777 & 0.866 & 0.832 & 0.702 & 0.829 & 0.792 & 0.733 & 0.868 & 0.747 & 0.853 & 0.935 & 0.840 & 0.796 & 0.907 & 0.820$\pm$0.002  & 0.344$\pm$0.003 & 0.864$\pm$0.002   \\
TA-DCL~\cite{zhang2023triplet}            & 0.773 & 0.885 & 0.825 & 0.699 & 0.826 & 0.788 & \textbf{0.734} & 0.877 & 0.744 & 0.849 & 0.932 & 0.829 & 0.795 & 0.935 & 0.821$\pm$0.002  & 0.350$\pm$0.002 & 0.868$\pm$0.001\\
MARL*~\cite{nie2023deep}                  & 0.777 & 0.869 & 0.833 & 0.703 & 0.831 & 0.794 & 0.732 & 0.870 & 0.750 & 0.853 & 0.936 & 0.838 & 0.798 & 0.907 & 0.821$\pm$0.001  & 0.352$\pm$0.001 & 0.872$\pm$0.002 \\
MAE~\cite{xiao2023delving}                & 0.770 & 0.876 & 0.824 & 0.718 & 0.828 & 0.771 & 0.732 & 0.880 & 0.745 & 0.862 & 0.940 & 0.831 & 0.785 & 0.940 & 0.822$\pm$0.002  & 0.356$\pm$0.002 & 0.872$\pm$0.002 \\
ThoraX-PriorNet*~\cite{hossain2023thorax} & 0.779 & 0.870 & 0.833 & 0.705 & 0.834 & 0.794 & 0.740 & 0.872 & 0.748 & 0.854 & 0.936 & 0.845 & \textbf{0.799} & 0.908 & 0.823$\pm$0.001  & 0.358$\pm$0.001 & 0.873$\pm$0.001 \\
KAD~\cite{zhang2023knowledge}             & \textbf{0.785} & \textbf{0.897} & 0.840 & 0.713 & 0.836 & 0.771 & 0.740 & 0.874 & 0.753 & 0.860 & 0.916 & 0.829 & 0.778 & \textbf{0.961} & 0.825            & 0.364   & 0.875  \\
\hline
ProtoPNet~\cite{chen2019looks}            & 0.761 & 0.861 & 0.810 & 0.701 & 0.820 & 0.783 & 0.712 & 0.872 & 0.743 & 0.836 & 0.936 & 0.801 & 0.780 & 0.912 & 0.809$\pm$0.002  & 0.326$\pm$0.002  & 0.830$\pm$0.003  \\
ProtoPNet++~\cite{wang2022Knowledge}      & 0.772 & 0.880 & 0.828 & 0.702 & 0.827 & 0.786 & 0.710 & 0.871 & 0.741 & 0.844 & 0.941 & 0.823 & 0.782 & 0.923 & 0.816$\pm$0.001  & 0.338$\pm$0.002  & 0.855$\pm$0.001  \\
Deformable~\cite{donnelly2022deformable}  & 0.769 & 0.878 & 0.834 & 0.706 & 0.824 & 0.790 & 0.707 & 0.872 & 0.744 & 0.837 & 0.943 & 0.820 & 0.785 & 0.929 & 0.817$\pm$0.001  & 0.341$\pm$0.001  & 0.859$\pm$0.002  \\
PIP-Net~\cite{nauta2023pip}               & 0.770 & 0.876 & 0.831 & 0.718 & 0.823 & 0.778 & 0.725 & 0.880 & 0.751 & 0.849 & 0.939 & 0.823 & 0.786 & 0.915 & 0.819$\pm$0.002  & 0.347$\pm$0.002  & 0.867$\pm$0.002  \\
XProtoNet~\cite{kim2021xprotonet}         & 0.780 & 0.887 & 0.835 & 0.710 & 0.831 & \textbf{0.804} & \textbf{0.734} & 0.871 & 0.747 & 0.840 & 0.941 & 0.815 & \textbf{0.799} & 0.909 & 0.822  & n/a & n/a  \\
CIPL                                      & 0.776 & 0.882 & 0.834 & \textbf{0.722} & 0.830 & 0.795 & \textbf{0.734} & \textbf{0.883} & 0.755 & 0.859 & \textbf{0.946} & \textbf{0.832} & 0.795 & 0.947 
& \textbf{0.828}$\pm$0.001  & \textbf{0.383}$\pm$0.002  & \textbf{0.885}$\pm$0.001  \\
\hline\hline
\end{tabular}

}
\vspace{-3pt}
\end{table*}

\begin{table*}[]
\caption{P-values computed from one-sided paired T-test between our CIPL and other competing approaches on NIH ChestX-ray14, evaluated with the mean AUC score.}
\vspace{-2pt}
\label{tab:T_test_xray}
\centering
\setlength{\tabcolsep}{0.83 mm}
\resizebox{1.0\linewidth}{!}{

\begin{tabular}{cccccccccc} 
\hline
ProtoPNet \cite{chen2019looks} 
& ProtoPNet++ \cite{wang2022Knowledge}   
& Deformable\cite{donnelly2022deformable} 
& PIP-Net \cite{nauta2023pip}   
& CPCL \cite{zhou2021multi} 
& TA-DCL \cite{zhang2023triplet} 
& MARL \cite{nie2023deep} 
& MAE \cite{xiao2023delving} 
& ThoraX-PriorNet \cite{hossain2023thorax} 
& KAD \cite{zhang2023knowledge} 
\\ 
\hline
4.9e-13            & 1.7e-7      & 1.9e-6     & 9.7e-6     & 3.8e-5      & 8.9e-5     & 1.1e-4     & 4.6e-4    &  9.0e-4     &  9.8e-3      \\
\hline
\end{tabular}

}
\vspace{-5pt}
\end{table*}

\vspace{-10pt}
\subsection{Experimental Settings}
\label{sec:settings}

Our proposed method adopts CNN-type feature backbone $f_{\theta_{f}}$ (DenseNet-121~\cite{huang2017densely} for chest X-rays and ResNet101~\cite{he2016deep} for fundus images), 
which are pre-trained on ImageNet, as in mainstream previous works.
To enable rigorous model evaluation, we utilise purely official train/test data provided by both datasets, without introducing any additional data or test-time augmentation, as suggested in~\cite{isensee2024nnu}.
Typically, input images are down-sampled 32 times by the backbone $f_{\theta_{f}}$, so $\bar{H} = \frac{H}{32}$ and $\bar{W} = \frac{W}{32}$.
We have the feature dimension $D=256$ by appending two $1 \times 1$ convolutional layers after $f_{\theta_{f}}$ (one activated by ReLU and the other activated by Sigmoid). 
For both tasks, we adopt $M=50$ prototypes per class, so $N=750$ and $400$ for X-rays and fundus images, respectively. 
Our CIPL is implemented with PyTorch platform and trained using the Adam optimiser with mini-batch sizes $|\mathcal{B}| = |\mathcal{B}^{\prime}| = 24$ and initial learning rate of $10^{-4}$, decreased by a rate of $0.5$ per $5$ epochs after reaching a steady performance. 
In Eq.~\eqref{eq:seperation}, we have the margin $\tau = 2$.
The weighting factors in Eq.~\eqref{eq:loss_basic} and Eq.~\eqref{eq:total_loss} are set as $\alpha_1=0.02$, $\alpha_2=\alpha_3=\alpha_4=0.5$. 
We first train our CIPL for 15 epochs using only $\mathcal{L}_{basic}$ in Eq.~\eqref{eq:loss_basic} to warm-up the model, 
then we use the overall objective $\mathcal{L}_{total}$ in Eq.~\eqref{eq:total_loss} to train the model for 40 epochs. 
Our method undergoes five runs with different random initialisation seeds, from which the results are reported. 

In Section~\ref{sec:intra}, our intra-image prototypical learning strategy relies on two augmented image views $\mathbf{x}_b$ and $\mathbf{x}_{b^{\prime}}$. 
However, the two views should not have spatially drastic disparities, 
since Eq.~(\ref{eq:loss_intraspatial}) enforces the interpretation alignment of similarity maps at the spatial level. 
To this end, we devise a two-step augmentation scheme. 
The first step applies basic location-oriented transformations, encompassing scaling, shearing, rotation, translation, and flipping, to increase the diversity of the training set. 
This augmented image serves as input to the next step, which applies colour-oriented transformations, involving changes in contrast, brightness, sharpness, and hue. 
To introduce even more variations between the two views, we also use a random crop of size 95\% to 100\% of the input, with aspect ratio between 0.95 and 1.05.
In this way, slight location variations are further incorporated between the two views, but pathological lesions at a specific location will roughly correspond to the same region in both views, and this can therefore help obtain consistent and robust prototypes. 
It is worthy noticing that the augmentations should preserve key semantic features of the images while avoiding excessive alterations which could distort the underlying class characteristics. 
Hence, we conduct a series of experiments by varying the strength of the used augmentations, 
where we apply the augmentations to the training images and monitor the model's performance on the test set. 
The optimal set of augmentations is determined based on the best observed performance. 
Our code will be made publicly available, from which more augmentation details can be found\footnote{https://github.com/cwangrun/CIPL}.


\begin{figure*}[!t]
    \centering
    \includegraphics[width=1.0\linewidth]{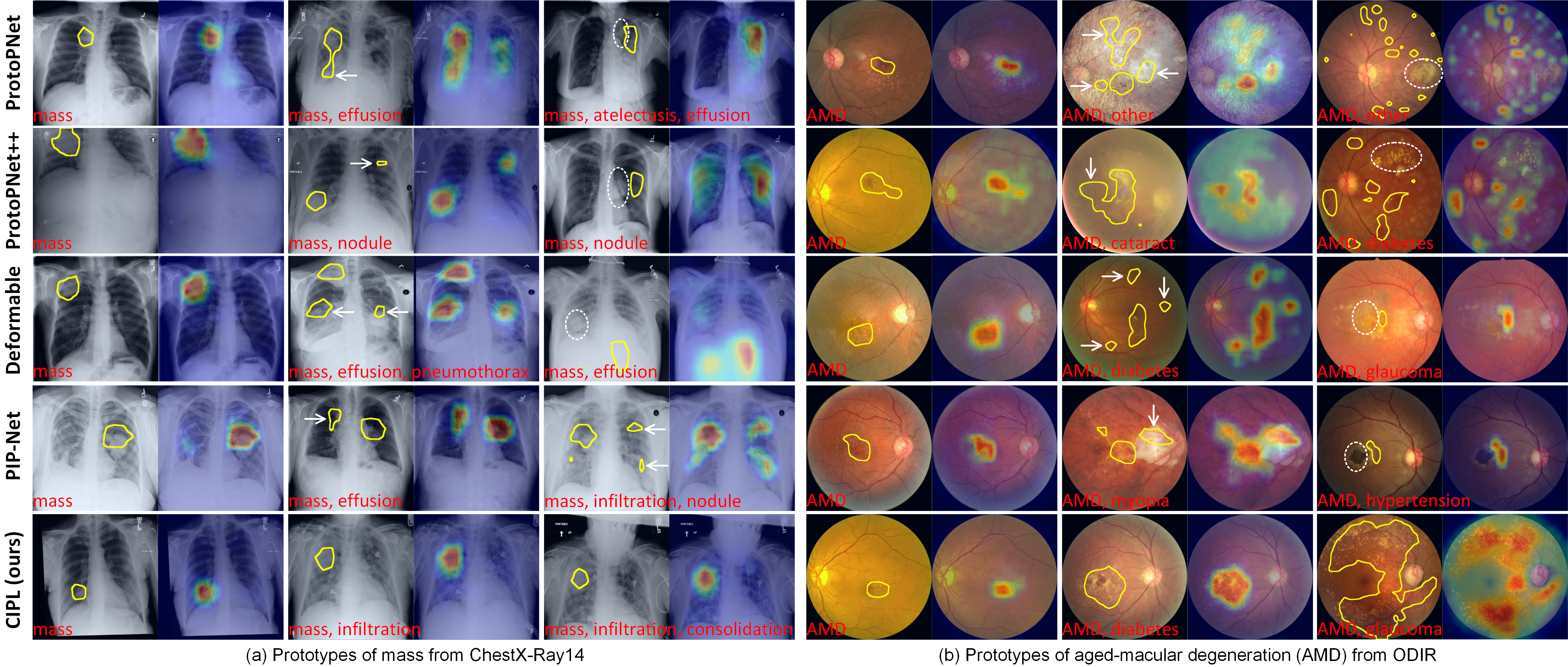}
    \vspace{-18pt}
    \caption{Visual comparison of prototypes learned from ProtoPNet~\cite{chen2019looks}, ProtoPNet++~\cite{wang2022Knowledge}, Deformable ProtoPNet~\cite{donnelly2022deformable}, PIP-Net~\cite{nauta2023pip}, and our CIPL. In each pair, the left-sided image displays the prototype (denoted by yellow counters) in the source training image and the corresponding right-sided image highlights prototypes in the self-activated similarity map. The ground-truth image labels are written in red.}
    \label{fig:prototypes}
    \vspace{-5pt}
\end{figure*}

\vspace{-8pt}
\subsection{Competing Methods}


\textit{Chest X-ray Classification:} 
we first compare with existing leading multi-label X-ray classification approaches~\cite{wang2017chestx,rajpurkar2017chexnet,guendel2019learning,ma2019multi,xiao2023delving,agu2021anaxnet,qi2022gren,ouyang2020learning,nie2023deep,hossain2023thorax,zhang2023triplet,zhang2022contrastive,wu2023medklip,zhang2023knowledge}, 
which are introduced in Section~\ref{sec:relatedwork_multilabel}. 
Furthermore, we also compare with state-of-the-art interpretable approaches that are based on prototypes primarily designed for single-label tasks: 
ProtoPNet~\cite{chen2019looks}, 
XProtoNet~\cite{kim2021xprotonet},
ProtoPNet++~\cite{wang2022Knowledge}, 
Deformable ProtoPNet~\cite{donnelly2022deformable}, and 
PIP-Net~\cite{nauta2023pip}.
ProtoPNet~\cite{chen2019looks} is the original work that utilises region-level prototypes for interpretable image classification, and severs as the baseline on which our CIPL is established. 
Relying on ProtoPNet, XProtoNet~\cite{kim2021xprotonet} further integrates prototypes with disease occurrence maps. 
ProtoPNet++~\cite{wang2022Knowledge} presents a two-branch approach ensembling the ProtoPNet with a non-interpretable global deep-learning classifier.
Deformable ProtoPNet~\cite{donnelly2022deformable} uses spatially-flexible prototypes to adaptively capture variations of object features from natural images. 
PIP-Net~\cite{nauta2023pip} first pre-trains class-agnostic prototypes and then regularises the prototype-class connection sparsity for compact classification interpretations.

\textit{Fundus Image Classification:} 
considering the limited research on multi-label ophthalmic disease diagnosis from fundus images, we begin by comparing our work with the following general multi-label image classification methods:
DB-Focal~\cite{wu2020distribution}, 
C-Tran~\cite{lanchantin2021general}, 
CPCL~\cite{zhou2021multi}, and 
TA-DCL~\cite{zhang2023triplet}, which are elaborated in Section~\ref{sec:relatedwork_multilabel}.
In addition, we also compare with the advanced prototype-based interpretable approaches introduced above: 
ProtoPNet~\cite{chen2019looks}, 
ProtoPNet++~\cite{wang2022Knowledge}, 
Deformable ProtoPNet~\cite{donnelly2022deformable}, and 
PIP-Net~\cite{nauta2023pip}.

\vspace{-8pt}
\subsection{Multi-label Disease Classification and Interpretation}

Table~\ref{tab:xrayscls} shows thoracic disease classification results of our CIPL and other competing approaches on NIH ChestX-ray14.
As can be observed, our CIPL method achieves the best classification results among all these approaches with a mean AUC of 0.828, a mean F1 score of 0.383, and a mean accuracy of 0.885, 
and it reaches the highest AUC results on five thoracic diseases.
Compared with the seminal work of ProtoPNet baseline~\cite{chen2019looks}, our CIPL exhibits a large average AUC improvement from 0.809 to 0.828. 
Notice that our method also greatly surpasses other more advanced interpretable prototypical learning approaches~\cite{wang2022Knowledge,donnelly2022deformable,nauta2023pip,kim2021xprotonet}, 
verifying the advantage of CIPL in exploiting cross- and intra-image information with prototypical learning for multi-label classification. 
Furthermore, our CIPL outperforms the current SOTA chest X-ray classification methods that rely on black-box deep neural networks, 
such as ThoraX-PriorNet~\cite{hossain2023thorax}, MARL~\cite{nie2023deep}, MAE~\cite{xiao2023delving}, and TA-DCL~\cite{zhang2023triplet}. 
Promisingly, our CIPL method exceeds the recent visual-language models (e.g., ConVIRT~\cite{zhang2022contrastive}, MedKLIP~\cite{wu2023medklip}, and KAD~\cite{zhang2023knowledge}), 
which are even pre-trained on image-text pairs from the large MIMIC-CXR~\cite{johnson2019mimic} database and further fine-tuned on ChestX-ray14.
These results highlight our method's strength of providing accurate diagnosis and interpretation results for thoracic diseases.
Table~\ref{tab:T_test_xray} provides a statistical analysis of our proposed CIPL with respect to other advanced competing approaches using the one-sided paired T-test. 
All these approaches undergo inference on ChestX-ray14 test set with 1000 bootstrap replicates. 
We can see that the performance improvements (in mean AUC) of our CIPL over the other considered approaches are statistically significant, with a significance level of 0.05.

\begin{figure}[!t] 
    \centering
    \includegraphics[width=1.0\linewidth]{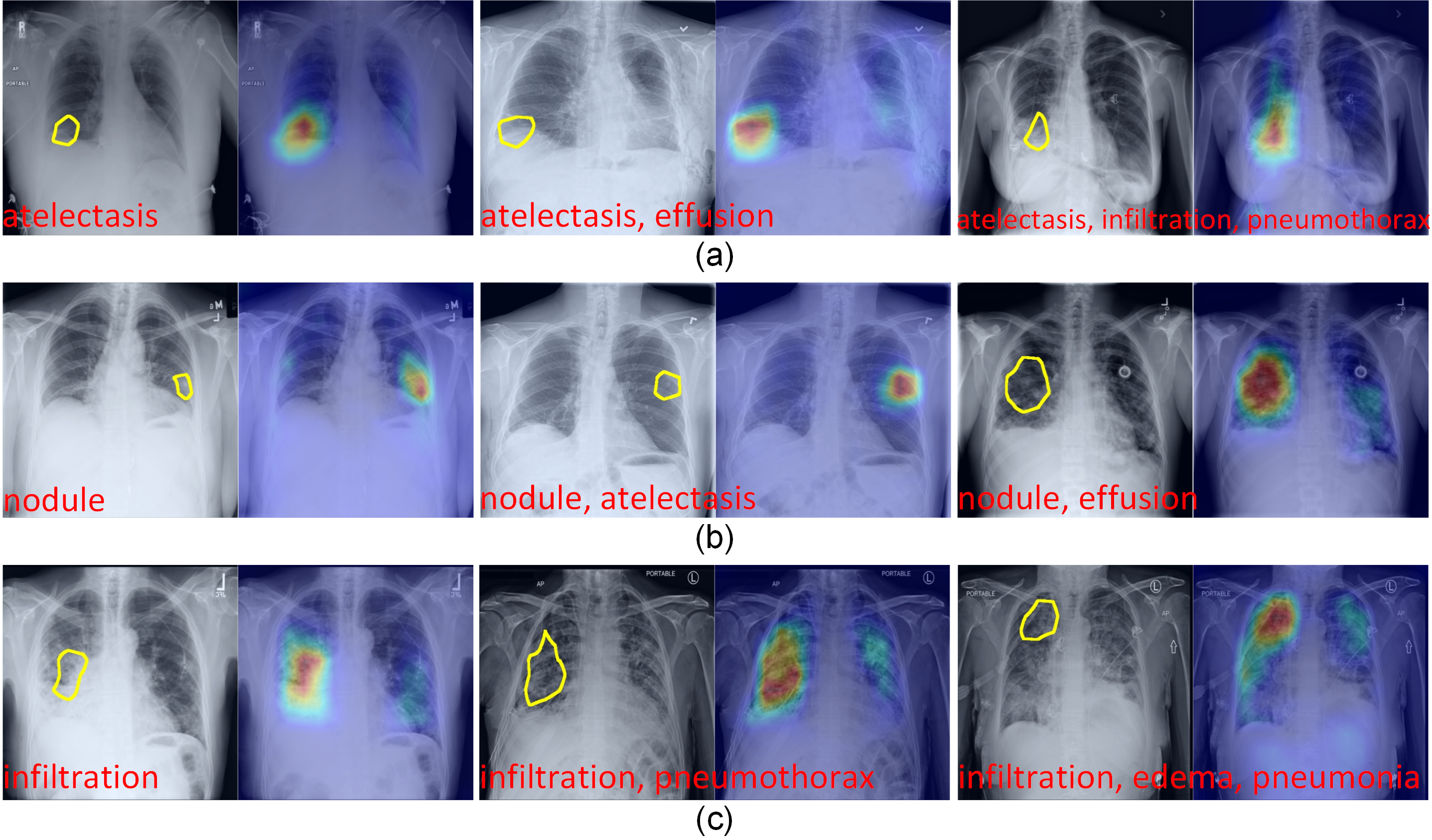}
    \vspace{-20pt}
    \caption{Visual prototypes of \textit{atelectasis} (a), \textit{nodule} (b), and \textit{infiltration} (c), learned by our CIPL method from NIH ChestX-ray14. In each pair, the left-sided image displays the prototype (denoted by yellow counters) in the source training image and the corresponding right-sided image highlights prototypes in the self-activated similarity map. The ground-truth image labels are written in red.}
    \label{fig:prototypes_more}
    \vspace{-6pt}
\end{figure}

Fig.~\ref{fig:prototypes}(a) provides a visual comparison of typical \textit{mass} prototypes learned by different prototype-based methods. 
It can be seen that, although obtaining good \textit{mass} prototypes from single-label training samples, 
the previous prototype-based methods can undesirably generate \textit{mass} prototypes that simultaneously activate \textit{mass} and other diseases 
(e.g., \textit{effusion} highlighted with the white arrow in the $2^{nd}$ case, at the top row), 
showing the phenomena of entangled prototypes. 
Also, the previous methods tend to produce inaccurate prototypes which are even not aligned with the actual \textit{mass} region (e.g., the white ellipse in the $3^{rd}$ case, at the top row), 
due to the disturbance of other diseases in multi-label training samples. 
By contrast, our CIPL successfully obtains disentangled and correct prototypes from both single- and multi-label samples.
We further provide visual prototypes of more other thoracic disease classes learned by our proposed CIPL method in Fig.~\ref{fig:prototypes_more},
showing their high disentanglement property. 

\begin{figure}[!t] 
    \centering
    \includegraphics[width=1.0\linewidth]{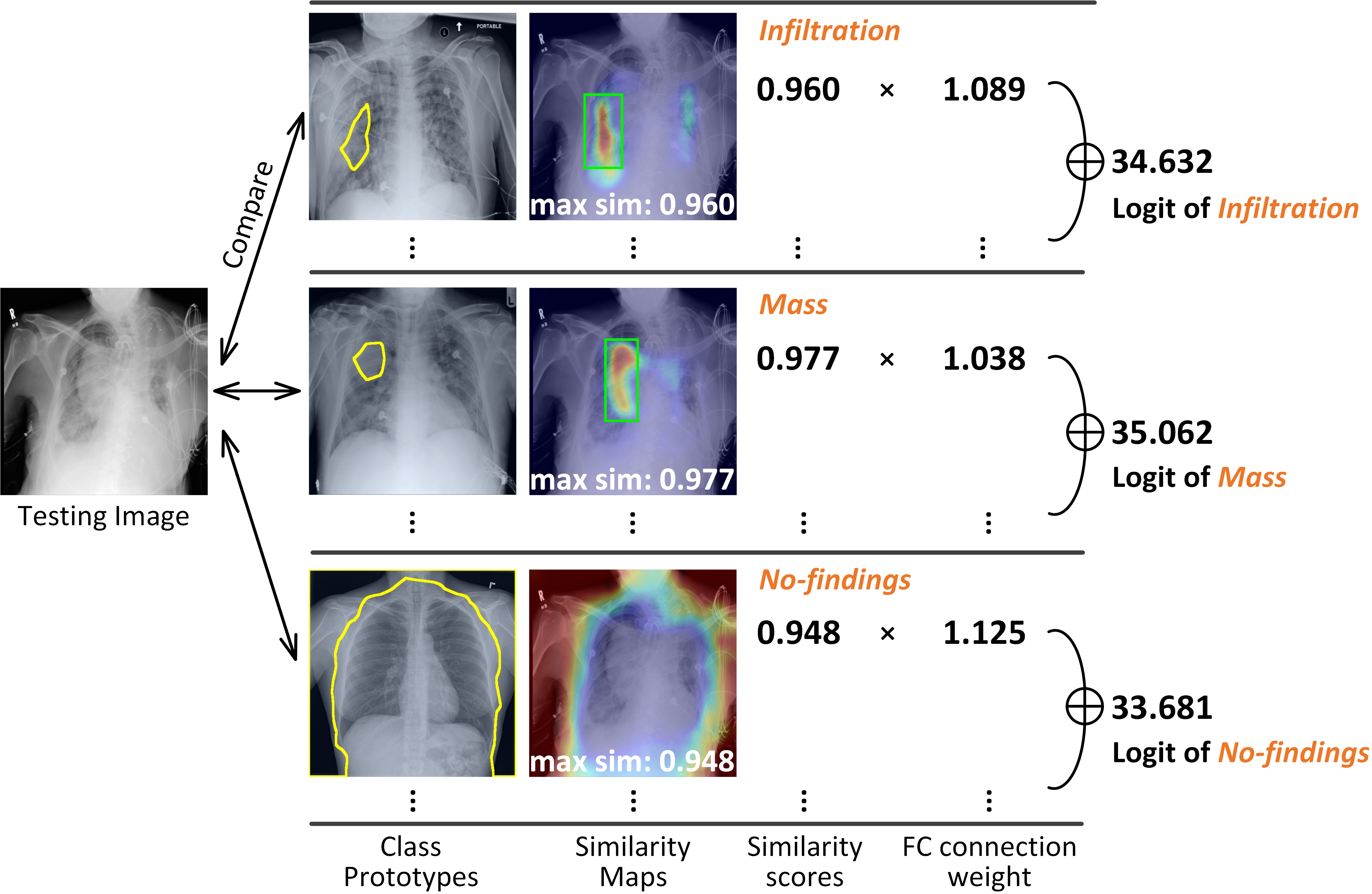}
    \vspace{-18pt}
    \caption{A typical example of our CIPL method for multiple disease diagnosis interpretation from a testing chest X-ray image, where the green boxes indicate ground-truth disease annotations. For simplicity, the figure only illustrates one prototype and the corresponding similarity map from the class \textit{infiltration}, \textit{mass}, and \textit{no-findings}. }
    \label{fig:reasoning}
    \vspace{-5 pt}
\end{figure}

In Fig.~\ref{fig:reasoning}, we display an example of multiple disease thoracic diagnosis interpretation from a testing chest X-ray image. 
As evident, our CIPL compares the input image with the learned class prototypes in the feature space, resulting in a series of similarity maps for each disease class and no-findings. 
The classification logits are generated by summing the weighted maximum similarity scores of the corresponding class. 
In this example, since the classification logits of \textit{infiltration} and \textit{mass} are higher than that of \textit{no-findings}, 
our model forms a disease diagnosis outcome of \textit{infiltration} and \textit{mass} from the image. 
It is worth noting that our model activates the corresponding disease regions that are well-aligned with the manual disease annotations from radiologist's domain knowledge.

\begin{table*}[]
\caption{Quantitative Results of Fundus Image Classification on ODIR Test Sets (On-site and Off-site), Reported with the AUC on the 7 Diseases, Mean AUC (mAUC), Mean F1 (mF1), and Mean Accuracy (mACC) Metrics. Top-part Shows General Classification Methods and Bottom Part Shows Prototype-based Approaches, with Best Results Highlighted in Bold. }
\label{tab:funduscls}
\vspace{-3pt}
\setlength{\tabcolsep}{0.6 mm}
\centering
\resizebox{1.0\linewidth}{!}{

\begin{tabular}{l|cccccccccc|cccccccccc} 
\hline\hline
\multirow{2}{*}{Method}              & \multicolumn{10}{c|}{On-site}                                                                                   & \multicolumn{10}{c}{Off-site}        \\ 
\cline{2-21}
                                     & Diab     & Glau     & Cata      & AMD      & Hype     & Myop     & Othe      & mAUC    & mF1   & mACC          & Diab      & Glau      & Cata     & AMD      & Hype     & Myop    & Othe    & mAUC    & mF1   & mACC  \\ 
\hline
DB-Focal~\cite{wu2020distribution}   & 0.819    & 0.857    & 0.921     & 0.860    & 0.794    & 0.913    & 0.547     & 0.816$\pm$0.004  & 0.463$\pm$0.005  & 0.819$\pm$0.006         & 0.869     & 0.823     & 0.952    & 0.874    & 0.801    & 0.913   & 0.630   & 0.837$\pm$0.005   & 0.543$\pm$0.005  & 0.868$\pm$0.004   \\
C-Tran~\cite{lanchantin2021general}  & 0.853 & \textbf{0.904} & 0.949  & 0.863    & 0.802    & 0.941    & 0.628     & 0.848$\pm$0.003  & 0.498$\pm$0.004  & 0.901$\pm$0.004         & 0.895     & 0.834     & 0.969    & 0.903    & 0.819    & 0.949   & 0.693   & 0.866$\pm$0.004   & 0.592$\pm$0.004  & 0.886$\pm$0.003   \\
CPCL~\cite{zhou2021multi}            & 0.871    & 0.896    & 0.967     & 0.870    & 0.803    & 0.929    & 0.651     & 0.855$\pm$0.003  & 0.503$\pm$0.003  & 0.895$\pm$0.003         & 0.907  & 0.839  & 0.972  & \textbf{0.923}   & 0.809    & 0.955   & 0.706   & 0.873$\pm$0.003   & 0.605$\pm$0.003  & 0.909$\pm$0.002   \\           
TA-DCL~\cite{zhang2023triplet}  & 0.877    & 0.873    & 0.971   & \textbf{0.884}   & 0.793   & 0.935    & 0.677     & 0.858$\pm$0.002  & 0.508$\pm$0.003  & 0.894$\pm$0.003         & 0.915     & 0.855     & 0.977    & 0.916    & 0.789    & 0.956   & 0.720   & 0.875$\pm$0.002   & 0.607$\pm$0.003  & 0.917$\pm$0.003   \\
\hline
ProtoPNet~\cite{chen2019looks}       & 0.871    & 0.826    & 0.951     & 0.846    & 0.820    & 0.951    & 0.562     & 0.832$\pm$0.004  & 0.473$\pm$0.004  & 0.889$\pm$0.004         & 0.910     & 0.812     & 0.952    & 0.898    & 0.851    & 0.957   & 0.657   & 0.862$\pm$0.004   & 0.565$\pm$0.004  & 0.880$\pm$0.004   \\
Deformable~\cite{donnelly2022deformable} & 0.879  & 0.834  & 0.976     & 0.862    & 0.835    & 0.958    & 0.565     & 0.844$\pm$0.004  & 0.481$\pm$0.003  & 0.891$\pm$0.003         & 0.905     & 0.827     & 0.963    & 0.907    & 0.865    & 0.951   & 0.665   & 0.869$\pm$0.004   & 0.573$\pm$0.003  & 0.897$\pm$0.003   \\
ProtoPNet++~\cite{wang2022Knowledge} & 0.881  & 0.801  & \textbf{0.989}  & 0.849  & 0.844    & 0.941    & 0.631     & 0.848$\pm$0.003  & 0.496$\pm$0.004  & 0.894$\pm$0.004         & 0.913     & 0.815     & 0.976    & 0.901    & 0.860    & 0.943   & 0.683   & 0.870$\pm$0.004   & 0.601$\pm$0.004  & 0.904$\pm$0.003   \\
PIP-Net~\cite{nauta2023pip}          & 0.882    & 0.850    & 0.941     & 0.852    & 0.843    & 0.942    & 0.650     & 0.851$\pm$0.003  & 0.498$\pm$0.003  & 0.896$\pm$0.002         & 0.903  & 0.828   & 0.953  & 0.900  & \textbf{0.885}    & 0.935   & 0.701   & 0.872$\pm$0.003   & 0.592$\pm$0.003  & 0.911$\pm$0.002   \\
CIPL  & \textbf{0.896}  & 0.888           & 0.984  & 0.864  & \textbf{0.867}  & \textbf{0.963}  & \textbf{0.696}  & \textbf{0.880}$\pm$0.002    & \textbf{0.553}$\pm$0.003   & \textbf{0.911}$\pm$0.002    
      & \textbf{0.930}  & \textbf{0.869}  & \textbf{0.990}  & 0.922  & 0.882    & \textbf{0.993}   & \textbf{0.738}    & \textbf{0.903}$\pm$0.002   & \textbf{0.620}$\pm$0.003   & \textbf{0.934}$\pm$0.002     \\
\hline\hline
\end{tabular}

}
\vspace{-6pt}
\end{table*}

Table~\ref{tab:funduscls} gives the quantitative results of ophthalmic disease diagnosis on ODIR dataset. 
It is observed that our CIPL consistently produces the best average classification AUC results on both test sets. 
Importantly, CIPL substantially exceeds existing prototype-based methods~\cite{chen2019looks,donnelly2022deformable,wang2022Knowledge,nauta2023pip}, 
by large margins of at least 2.9\% and 3.1\% (in mean AUC) on the on-site and off-site test sets, respectively, 
evidencing the effectiveness of cross- and intra-image prototypical learning for multi-label fundus image classification. 
We showcase typical visual prototypes of the \textit{AMD} disease of different methods in Fig.~\ref{fig:prototypes}(b). 
Our CIPL generates AMD prototypes that precisely cover the pathological lesions associated with the \textit{AMD} disease, 
whereas the AMD prototypes from other compared methods are prone to be entangled with other diseases (see the white arrows), 
and in some cases, they completely neglect the actual \textit{AMD} lesions (indicated by the white ellipses). 

\vspace{-7pt}
\subsection{Weakly-supervised Multi-label Disease Localisation}

We also evaluate the effectiveness of our CIPL method for localising thoracic diseases in a weakly-supervised setting on NIH ChestX-ray14, 
where the average class activation map is leveraged in the localisation process elaborated in Section~\ref{sec:dataset}. 
In this task, we compare with current SOTA weakly-supervised localisation approaches that are based on CAM~\cite{wang2017chestx,hermoza2020region} and 
visual attentions~\cite{cai2018iterative,liu2019align,ouyang2020learning}. 
Table~\ref{tab:xraysloc} presents the thresholded IoU accuracy results, 
where the types of explanation used for disease localisation are also provided. 
As can be seen, our CIPL demonstrates a significant enhancement in localisation accuracy across most diseases, at IoU thresholds of both 0.1 and 0.3, 
achieving the average accuracy of 0.78 and 0.47, respectively. 
Compared with previous leading saliency-based (e.g., CAM and visual attentions) localisation methods, our CIPL exhibits a much higher accuracy.   
In particular, CIPL also yields superior results over other prototype-based methods~\cite{wang2022Knowledge,donnelly2022deformable,nauta2023pip} that consider only individual-image information in prototype's learning. 
These improvements are attributed to the utilisation of 
rich cross-image semantics to accurately understand the pathological lesions of each disease class 
and beneficial intra-image cues to produce consistent and robust prototypes, 
evidenced by exhaustive ablation studies in the following Section~\ref{sec:ablation}.


\begin{table}[]
\caption{Weakly-supervised Thoracic Disease Localisation Accuracy from Different Explanation Methods: Class Activation Map (CAM), Attention Map (Atten), and Prototype (Proto). }
\label{tab:xraysloc}
\vspace{-3pt}
\setlength{\tabcolsep}{0.50 mm}
\resizebox{\linewidth}{!}{

\begin{tabular}{l|c|c|ccccccccc} 
\hline\hline
\makebox[0.0\textwidth][l]{Method}   &\makebox[0.038\textwidth][c]{\textit{T}(IoU)}     & Type   & Atel  & Card  & Effu  & Infi  & Mass  & Nodu  & Pnea  & Pnex  & Mean   \\ 
\hline
Wang \textit{et al.}~\cite{wang2017chestx}  &\multirow{7}{*}{0.1}      & CAM      & 0.69  & 0.94  & 0.66  & 0.71  & 0.40  & 0.14  & 0.63  & 0.38  & 0.57   \\
Liu \textit{et al.}~\cite{liu2019align}          &                     & Atten    & 0.39  & 0.90  & 0.65  & 0.85  & 0.69  & 0.38  & 0.30  & 0.39  & 0.60   \\
Ouyang \textit{et al.}~\cite{ouyang2020learning}  &                    & Atten    & \textbf{0.78}  & 0.97  & \textbf{0.82}  & 0.85  & 0.78  & 0.56  & 0.76  & 0.48  & 0.75   \\
ProtoPNet~\cite{chen2019looks}                     &     &\multirow{4}{*}{Proto}  & 0.63  & 0.89  & 0.63  & 0.88  & 0.68  & 0.55  & 0.82  & 0.39  & 0.68   \\
ProtoPNet++~\cite{wang2022Knowledge}                &                  &          & 0.67  & 0.90  & 0.65  & 0.92  & 0.74  & 0.58  & 0.90  & 0.39  & 0.72   \\
Deformable~\cite{donnelly2022deformable}            &                  &          & 0.70  & 0.92  & 0.65  & \textbf{0.94}  & 0.71  & \textbf{0.59}  & 0.92  & 0.45  & 0.74   \\
CIPL                                                &                  &          & 0.69  & \textbf{1.00}  & 0.71  & 0.91  & \textbf{0.85} & 0.58  & \textbf{0.98}  & \textbf{0.48}  & \textbf{0.78}  \\ 
\hline
Wang \textit{et al.}~\cite{wang2017chestx}   &\multirow{9}{*}{0.3}     & CAM     & 0.24  & 0.46  & 0.30  & 0.28  & 0.15  & 0.04  & 0.17  & 0.13  & 0.22   \\
Cai \textit{et al.}~\cite{cai2018iterative}   &                        & Atten   & 0.33  & 0.85  & 0.34  & 0.28  & 0.33  & 0.11  & 0.39  & 0.16  & 0.35   \\
Ouyang \textit{et al.}~\cite{ouyang2020learning}  &                    & Atten   & 0.34  & 0.40  & 0.27  & 0.55  & \textbf{0.51}  & 0.14  & 0.42  & \textbf{0.22}  & 0.36   \\
Liu \textit{et al.}~\cite{liu2019align}        &                       & Atten   & 0.34  & 0.71  & \textbf{0.39}  & \textbf{0.65}  & 0.48  & 0.09  & 0.16  & 0.20  & 0.38   \\
Hermoza \textit{et al.}~\cite{hermoza2020region}  &                    & CAM     & \textbf{0.37}  & 0.99  & 0.37  & 0.54  & 0.35  & 0.04  & 0.60  & 0.21  & 0.43   \\
ProtoPNet~\cite{chen2019looks}               &        &\multirow{4}{*}{Proto}    & 0.29  & 0.83  & 0.21  & 0.42  & 0.37  & 0.19  & 0.50  & 0.14  & 0.37   \\
ProtoPNet++~\cite{wang2022Knowledge}         &                            &      & 0.31  & 0.86  & 0.28  & 0.51  & 0.39  & 0.20  & 0.52  & 0.13  & 0.40   \\
Deformable~\cite{donnelly2022deformable}      &                           &      & 0.33  & 0.84  & 0.30  & 0.54  & 0.41  & 0.19  & 0.54  & 0.16  & 0.41   \\
CIPL                                          &                           &      & 0.33  & \textbf{1.00}  & 0.26  & 0.53  & 0.47  & \textbf{0.33}  & \textbf{0.64}  & 0.18  & \textbf{0.47}   \\ 
\hline\hline
\end{tabular}

}
\vspace{-5pt}
\end{table}

\vspace{-3.3pt}
\subsection{Ablation Studies}
\label{sec:ablation}

\subsubsection{Cross-image Prototypical Learning}
We perform ablation experiments to study the effectiveness of our cross-image strategy in improving the disease classification and localisation performances.
Table~\ref{tab:ablation} provides the results on ODIR and ChestX-ray14,
where the baseline CIPL only uses the basic training objective $\mathcal{L}_{basic}$, which is the same as the ProtoPNet~\cite{chen2019looks} method. 
We notice that the cross-image prototypical learning strategy, denoted by $\mathcal{L}_{cross}$, can significantly improve the classification performance on both tasks. 
Also, this strategy greatly contributes to more accurate thoracic disease localisation, 
since it exploits common semantics of paired images to comprehensively understand the pathological lesions.

\begin{table}[]
\caption{Ablation Analysis of Our Proposed CIPL method. We Give Average Classification AUC Results on ODIR (On-site) and ChestX-ray14, and Average Thoracic Disease Localisation Accuracy (Loc. Acc.) at 0.3 IoU Threshold. }
\label{tab:ablation}
\vspace{-3pt}
\setlength{\tabcolsep}{1.6 mm}
\resizebox{\linewidth}{!}{

\begin{tabular}{cccccccc} 
\hline\hline
\multirow{2}{*}{$\mathcal{L}_{basic}$} & \multirow{2}{*}{$\mathcal{L}_{cross}$} & \multirow{2}{*}{$\mathcal{L}_{inte}$} & \multirow{2}{*}{${\mathcal{L}_{pred}}^\dagger$} & \multirow{2}{*}{$\mathcal{L}_{pred}$} & \multirow{2}{*}{\begin{tabular}[c]{@{}c@{}}ODIR\\(AUC)\end{tabular}} & \multicolumn{2}{c}{ChestX-ray14}  \\
                          &                            &                            &                             &                          &                     & AUC               & Loc. Acc.        \\ 
\hline 
\checkmark                &                            &                            &                             &                          & 0.832               & 0.809             & 0.37            \\
\checkmark                & \checkmark                 &                            &                             &                          & 0.863               & 0.824             & 0.43            \\
\checkmark                &                            & \checkmark                 &                             &                          & 0.841               & 0.814             & 0.44            \\
\checkmark                &                            &                            & \checkmark                  &                          & 0.836               & 0.811             & 0.39            \\
\checkmark                &                            &                            &                             &  \checkmark              & 0.856               & 0.820             & 0.40            \\
\checkmark                & \checkmark                 & \checkmark                 &                             &  \checkmark              & \textbf{0.880}      & \textbf{0.828}    & \textbf{0.47}   \\
\hline\hline
\end{tabular}

}
\end{table}

\subsubsection{Intra-image Prototypical Learning}

In Table~\ref{tab:ablation}, we also provide ablation analysis for our intra-image prototypical learning strategy. 
Regarding the prediction alignment, one may have concerns about simply aligning predictions of the two views for each individual training sample, mentioned in Section~\ref{sec:intra}. 
Hence, we also provide such results by minimising the KL-divergence of predictions from two views of individual samples, denoted by ${\mathcal{L}_{pred}}^{\dagger}$.
It can be noticed that, 
the interpretation alignment $\mathcal{L}_{inte}$ leads to significantly improved accuracy in the localisation of diseases 
since it is designed to make prototypes robust to two augmented image views. 
By contrast, the batch-level prediction alignment $\mathcal{L}_{pred}$ exhibits substantial gains in AUC classification performances, 
outperforming the sample-level prediction alignment ${\mathcal{L}_{pred}}^{\dagger}$, 
since $\mathcal{L}_{pred}$ considers holistic relationships among samples to provide more useful regularisation signals for the model.

\subsubsection{Prototype Number}

\begin{figure}[!t]
    \centering
    \includegraphics[width=0.85\linewidth]{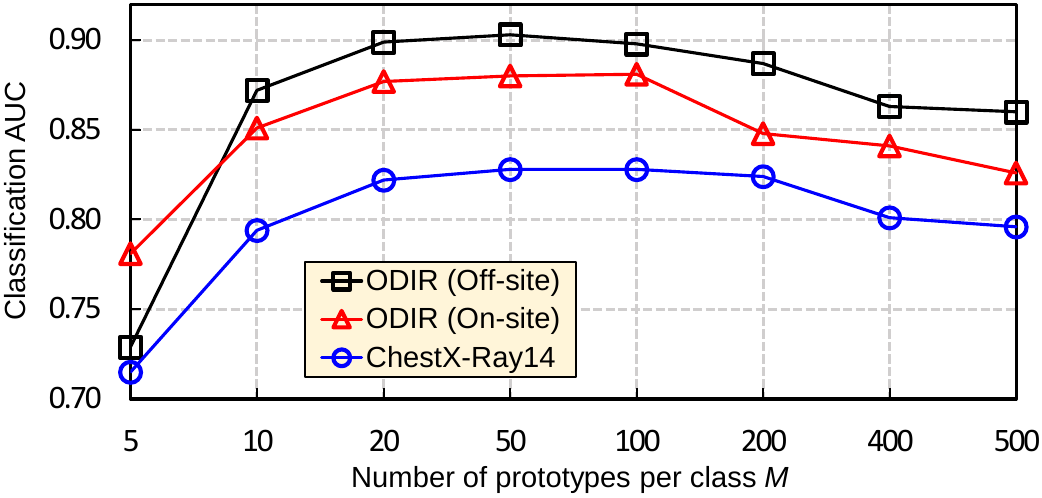}
    \vspace{-8pt}
    \caption{Sensitivity of our CIPL method to the number of prototypes per class, measured by the average classification AUC performance. }
    \label{fig:prototypenumber}
    \vspace{-3pt}
\end{figure}

We study the sensitivity of our CIPL method to the number of prototypes per class $M$, as depicted in Fig.~\ref{fig:prototypenumber}.
As evident, the classification results generally exhibit robustness to $M$ from 20 to 100. 
Nonetheless, a small number of prototypes leads to inadequate representation of rich and diverse patterns in training samples and results in sub-optimal results.
Conversely, an excessive number of prototypes introduces unnecessary and even harmful information, 
causing a decline in performance and an escalation in model complexity. 
According to Fig.~\ref{fig:prototypenumber}, we set the number of prototypes at 50 per class in all remaining experiments.

\vspace{-8pt}
\subsection{Single-label Diagnosis Scenario} 
\label{sec:single-label}

Our CIPL method is developed to tackle the multi-label disease diagnosis and interpretation,
but it is necessary to explore its effectiveness 
on the single-label diagnosis protocol. 
To achieve this, we adopt the chest X-ray dataset publicly accessible in~\cite{kermany2018identifying}. 
The dataset comprises 5,856 paediatric chest X-ray images, with 5,232 samples for training and 624 samples for testing. 
These images are categorised into one of the following three classes: normal, bacterial pneumonia, and viral pneumonia. 
In this experiment, we employ the same training details (batch size, learning rate, per-class prototype number), 
hyper-parameters ($\alpha_1, \alpha_2, \alpha_3, \alpha_4$ and $\tau$), and data augmentations, as elaborated in Section~\ref{sec:settings}. 

Table~\ref{tab:SinLabelCls} provides experimental results for the single-label chest X-ray classification task, 
where $\mathcal{L}_{intra} = \mathcal{L}_{inte} + \mathcal{L}_{pred}$, and we evaluate the overall sensitivity, specificity, precision, accuracy, and AUC score.  
As it can be observed, the cross-image training strategy $\mathcal{L}_{cross}$ enables slightly improved performances, 
compared with baseline CIPL only using $\mathcal{L}_{basic}$. 
This is due to only one type of disease in single-label training images, 
from which the learned prototypes are naturally disentangled. 
Combining with ablation results from Table~\ref{tab:ablation}, 
it can be affirmed that our cross-image strategy is effective in tackling entangled multi-label disease diagnosis. 

In Table~\ref{tab:SinLabelCls}, our proposed intra-image learning paradigm $\mathcal{L}_{intra}$ demonstrates notable performance improvements, 
since it leverages consistent and meaningful intra-image information in a self-supervised manner, regardless of whether the training images have single or multiple labels.

\begin{table}[]
\caption{Classification Results of Our Proposed CIPL Method from Single-label Pediatric Chest X-ray Images.}
\label{tab:SinLabelCls}
\setlength{\tabcolsep}{0.5 mm}
\resizebox{\linewidth}{!}{

\begin{tabular}{lccccc} 
\hline\hline
Method                                                                             & Sensitivity & Specificity & Precision & Accuracy       & AUC             \\   \hline
PIP-Net~\cite{nauta2023pip}                                                        & 0.894       & 0.949       & 0.890     & 0.897          & 0.965           \\
CIPL ($\mathcal{L}_{basic}$)                                                       & 0.882       & 0.943       & 0.887     & 0.891          & 0.960           \\
CIPL ($\mathcal{L}_{basic}$, $\mathcal{L}_{cross}$)                                & 0.887       & 0.950       & 0.898     & 0.898          & 0.963           \\
CIPL ($\mathcal{L}_{basic}$, $\mathcal{L}_{intra}$)                                & 0.914       & 0.961       & 0.925     & 0.926          & 0.975           \\
CIPL ($\mathcal{L}_{basic}$, $\mathcal{L}_{cross}, \mathcal{L}_{intra}$)           & \textbf{0.918}  & \textbf{0.964}  & \textbf{0.930}  & \textbf{0.931}  & \textbf{0.976}        \\
\hline\hline
\end{tabular}

}
\vspace{-5pt}
\end{table}

\vspace{6pt}
\section{Discussion}
\label{sec:discussion}

In this paper, we introduce the CIPL framework, designed to effectively learn disentangled prototypes for multi-label disease diagnosis. 
Our CIPL framework provides compelling similarity-based interpretations for the diagnosis decision with train-test associations, 
i.e, comparing testing samples with class-specific prototypes that represent particular regions of training images. 
In addition to relying on the region-level prototypes, the similarity-based interpretation can also be realised by image-level prototypes~\cite{li2018deep,wang2023visual,hesse2024prototype}, 
which are produced through different learning techniques such as clustering~\cite{wang2023visual} and auto-encoder models~\cite{li2018deep,gautam2022protovae}. 
However, these image-level prototypes are based on whole-image representations, resulting in less fine-grained interpretability compared to the region-level prototypes employed in our method. 
From a human user's perspective, understanding which parts of a testing image (e.g., pathological lesions in medical images) are similar to the region-level prototypes is more meaningful than merely determining whether the entire testing sample resembles the whole-image training prototypes. 

Prototypical learning aims to find a set of class-specific prototypes to describe or summarise the training data, 
with inference relying on the similarities between testing samples and these prototypes.  
In this sense, the prototypical learning can be broadly categorised as a form of transductive reasoning~\cite{mao2021joint}, 
where distances or affinities to labelled reference samples are leveraged for the classification process. 
Even if the procedure is straightforward and well-understood by humans, 
its effectiveness often hinges on the complexity of the training data's distribution as well as the prototype's capacity of approximating such distribution~\cite{wang2023mixture}. 
This issue is clarified in Fig.~\ref{fig:prototypenumber}, revealing that both an excessive or insufficient number of prototypes can negatively impact model performance.

\vspace{2pt}

\begin{figure}[t!]
    \centering
    \includegraphics[width=1.00\linewidth]{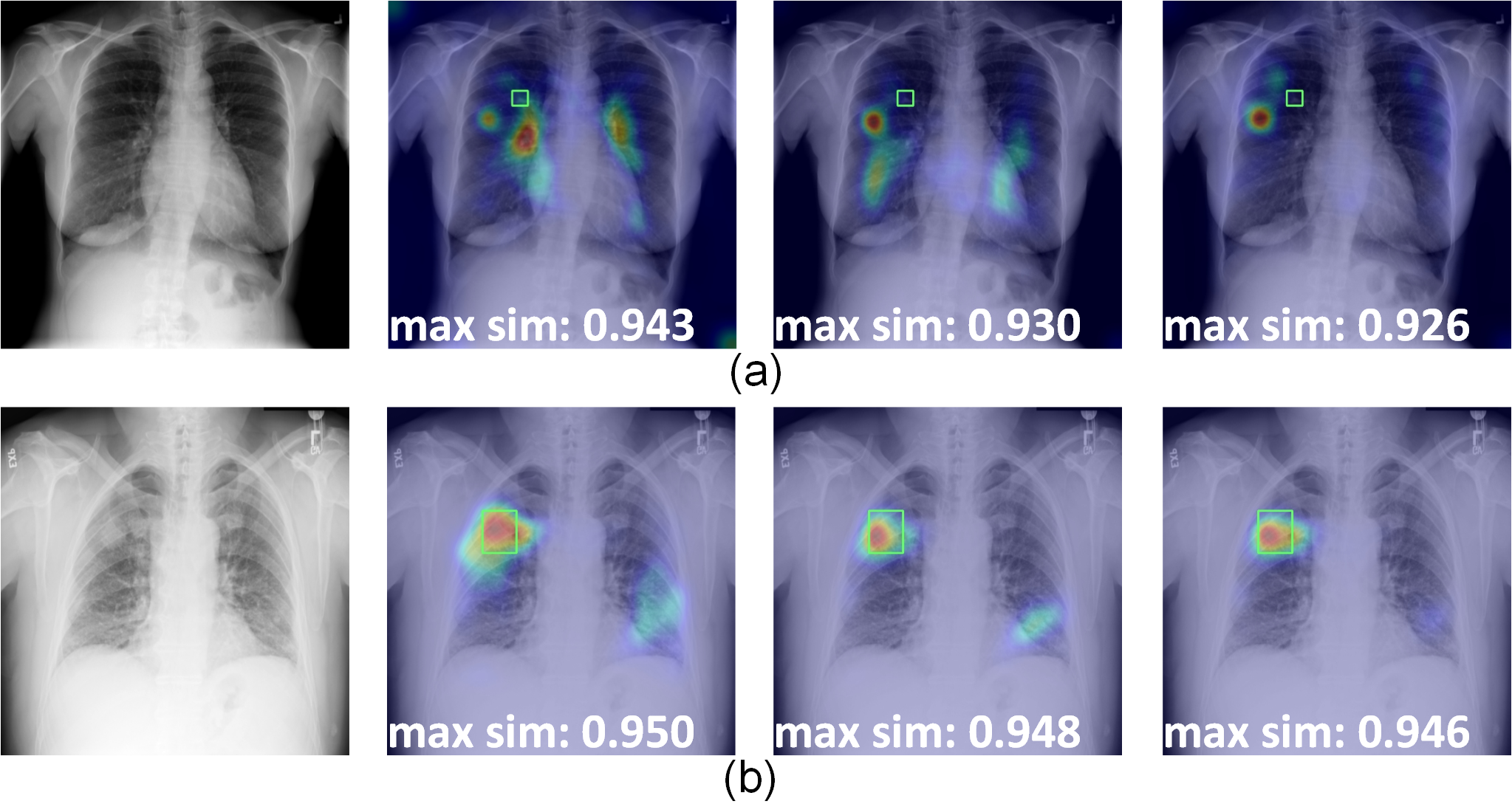}
    \vspace{-19pt}
    \caption{Failure examples of our CIPL method. 
    (a) mis-localisation of the subtle \textit{nodule} lesion, where we show top-3 similarity maps with the true \textit{nodule} class. 
    (b) mis-classification for a \textit{nodule} image, where we show top-3 similarity maps with the false predicted \textit{mass} class. 
    The green box in each similarity map denotes ground-truth \textit{nodule} annotations. 
    }
    \label{fig:failure} 
    \vspace{-5pt}
\end{figure}

While the prototype-based classification approach yields appealing interpretable results for disease diagnosis, 
our experiments still reveal some unfavourable failure cases that fall into two main types. 
1) Mis-localisation of subtle pathological lesions, as shown in Fig.~\ref{fig:failure}(a), 
occurs when the lesions (such as \textit{nodule} highlighted by the green box) are extremely small, 
making them difficult to differentiate from the normal tissues and leading to the prototype activation in incorrect areas. 
It is noteworthy that the mis-localisation does not always cause wrong classification. 
Because of the model's transparency, such cases are easily identified, allowing for further model refinement.
2) Mis-classification to visually similar classes, as illustrated in Fig.~\ref{fig:failure}(b), 
where the \textit{mass} prototypes also activate the \textit{nodule} lesion, resulting in a false positive prediction of the \textit{mass} class. 
We believe that this is partly because both the \textit{nodule} and \textit{mass} have restricted and regular shape, which are poorly-distinguishable between them in the image. 
To address this, in the future we plan to incorporate constraints during training to encourage the prototype separability for these visually similar disease classes. 

\section{Conclusion}
\vspace{-2pt}
\label{sec:conclusion}

In this work, we presented the CIPL method for multi-label disease diagnosis and interpretation from medical images. 
Our method leverages common cross-image semantics to disentangle the disease features of multiple classes, 
allowing for a more effective learning of class prototypes to comprehensively understand various pathological lesions. 
Moreover, our method exploits consistent intra-image information to enhance interpretation robustness and predictive performance, 
through an alignment-based regularisation strategy. 
Experimental results on two multi-label benchmarks illustrate the advantages of our CIPL over other competing approaches in disease classification and interpretation (measured by weakly-supervised localisation). 
Furthermore, our intra-image prototypical learning strategy is versatile and easily extended to tackle the single-label disease diagnosis and interpretation.



\bibliographystyle{ieeetr}
\bibliography{refs.bib}

\end{document}